\documentclass[10pt, letter, onecolumn]{main}

\usepackage[pdftex]{graphicx}
\usepackage{pifont}
\usepackage{amssymb}
\usepackage{multirow}
\usepackage{array}
\usepackage{dblfloatfix}
\usepackage{titlesec}
\usepackage[numbers,sort&compress]{natbib}
\usepackage{nameref}
\usepackage{varioref}
\usepackage{etoc}

\newlength{\cboxlength}
\usepackage{amsmath,xparse,mleftright}
\NewDocumentCommand{\up}{som}{%
  \IfBooleanTF{#1}
    {\upext{#3}}
    {#3\IfNoValueTF{#2}{\mathord}{#2}\uparrow}%
}
\NewDocumentCommand{\upext}{m}{%
  \mleft.\kern-\nulldelimiterspace#1\mright\uparrow
}

\usepackage{xcolor} 
\usepackage{soul} 

\usepackage{adjustbox}
\usepackage[most]{tcolorbox}
\usepackage{float}
\usepackage{xspace}
\usepackage[symbol]{footmisc}
\usepackage{lineno}
\usepackage{rotating}%
\usepackage{booktabs}%
\usepackage{makecell}%

\tcbset{
  aibox/.style={
    width=394.18663pt,
    top=10pt,
    colback=white,
    colframe=black,
    colbacktitle=black,
    enhanced,
    center,
    attach boxed title to top left={yshift=-0.1in,xshift=0.15in},
    boxed title style={boxrule=0pt,colframe=white,},
  }
}
\newtcolorbox{AIbox}[2][]{aibox,title=#2,#1}

\ifdefined\red
    \renewcommand{\red}[1]{\textcolor{red}{#1}}
\else
    \newcommand{\red}[1]{\textcolor{red}{#1}}
\fi
\ifdefined\blue
    \renewcommand{\blue}[1]{\textcolor{blue}{#1}}
\else
    \newcommand{\blue}[1]{\textcolor{blue}{#1}}
\fi

\newcommand{\modelname}{\mbox{RoentGen-v2 }}

\makeatletter
\renewcommand\AB@authnote[1]{}
\renewcommand\AB@affilnote[1]{}
\makeatother

\usepackage{anyfontsize} 
\usepackage{titlesec}
\titleformat{\section}{\normalfont\Large\bfseries}{\thesection}{1em}{#1}

\title{{\fontsize{16.5pt}{15.5pt}\selectfont Improving Performance, Robustness, and Fairness of Radiographic AI Models with Finely-Controllable Synthetic Data}}

\author[]{
  Stefania L. Moroianu$^{1,3,*}$,
  Christian Bluethgen$^{1,5}$,
  Pierre Chambon$^{1}$,
  Mehdi Cherti$^{6,7}$,
  Jean-Benoit Delbrouck$^{1,2}$,
  Magdalini Paschali$^{1,2}$,
  Brandon Price$^{8,9}$,
  Judy Gichoya$^{8}$,
  Jenia Jitsev$^{6,7}$,\\
  Curtis P. Langlotz$^{1,2}$,
  Akshay S. Chaudhari$^{1,2,4}$
}

\affil{\footnotesize{
$^{1}$ Center for Artificial Intelligence in Medicine and Imaging, Stanford University. 
$^{2}$ Department of Radiology, Stanford University.
$^{3}$ Department of Applied Physics, Stanford University.
$^{4}$ Department of Biomedical Data Science, Stanford University.
$^{5}$ Institute for Diagnostic and Interventional Radiology, University Hospital Zurich, University of Zurich.
$^{6}$ LAION.
$^{7}$ Juelich Supercomputing Center (JSC), Research Center Juelich (FZJ).
$^{8}$ Department of Radiology \& Imaging Sciences, Emory University.
$^{9}$ Department of Radiology, University of Florida College of Medicine.
}}

\renewcommand{\correspondingauthor}[1]{$\ast$~Corresponding author: slmoro@stanford.edu}

\usepackage{xcolor}
\usepackage{hyperref}

\hypersetup{
    colorlinks=true,   
    urlcolor=blue      
}
\usepackage[noabbrev,capitalize]{cleveref}

\begin{document}
\begin{abstract}
Achieving robust performance and fairness across diverse patient populations remains a central challenge in developing clinically deployable deep learning models for diagnostic imaging. 
Synthetic data generation has emerged as a promising strategy to address current limitations in dataset scale and diversity. 
In this study, we introduce \modelname, a state-of-the-art text-to-image diffusion model for chest radiographs that enables fine-grained control over both radiographic findings and patient demographic attributes, including sex, age, and race/ethnicity. 
\modelname is the first model to generate clinically plausible chest radiographs with explicit demographic conditioning, facilitating the creation of a large, demographically balanced synthetic dataset comprising over 565,000 images.
We use this large synthetic dataset to evaluate optimal training pipelines for downstream disease classification models. 
In contrast to prior work that combines real and synthetic data naively, we propose an improved training strategy that leverages synthetic data for supervised pretraining, followed by fine-tuning on real data. 
Through extensive evaluation on over 137,000 held-out chest radiographs from five institutions, we demonstrate that synthetic pretraining consistently improves model performance, generalization to out-of-distribution settings, and fairness across demographic subgroups defined across varying fairness metrics.
Across datasets, synthetic pretraining led to a 6.5\% accuracy increase in the performance of downstream classification models, compared to a modest 2.7\% increase when naively combining real and synthetic data. We observe this performance improvement simultaneously with the reduction of the underdiagnosis fairness gap by 19.3\%, with marked improvements across intersectional subgroups of sex, age, and race/ethnicity.
Our proposed data-centric training approach that combines high-fidelity synthetic training data with multi-stage training pipelines is label-efficient, reducing reliance on large quantities of annotated real data. 
These results highlight the potential of demographically controllable synthetic imaging to advance equitable and generalizable medical deep learning under real-world data constraints. We open source our code, trained models, and synthetic dataset at \href{https://github.com/StanfordMIMI/RoentGen-v2}{https://github.com/StanfordMIMI/RoentGen-v2}.
\end{abstract}

\maketitle

\vspace{10mm}

\nolinenumbers
\clearpage
\section{Introduction}\label{sec1}
Chest radiography (CXR) is the most frequently performed medical imaging exam worldwide, playing a central role in both diagnostic workflows and routine screening.
An estimated 2 billion CXR exams are acquired globally each year~\cite{Akhter2023}, including over 70 million in the United States~\cite{Iyeke2022}. 
Given this large volume of CXR studies and the global shortage of radiologists, deep learning models have emerged as promising tools to support radiographic interpretation~\cite{all2021}. 
Such models have demonstrated strong performance across a wide range of tasks, including disease classification~\cite{Tiu2022,AitNasser2023,Kim2023,Hsieh2023}, worklist prioritization~\cite{Baltruschat2020}, visual grounding~\cite{BiovilBoecking2022,BiovilT}, and automatic report generation~\cite{cxrReportGen,chexagent,Tanno2024}, 
with many systems obtaining regulatory clearances~\cite{Milam2023}. 
Despite these initial successes, several shortcomings remain that limit the translation of deep learning models into routine clinical practice.
A major concern is the lack of generalizability: models trained on data from one institution often exhibit degraded performance when applied to external sites or new patient populations~\cite{generalizationCohen,FernndezMiranda2024}.
Moreover, fairness concerns arise when models exhibit performance discrepancies across demographic subgroups, potentially contributing to inequitable clinical care~\cite{chexclusion,NatureSeyyedKalantari2021,Ahluwalia2023}. 
Although fairness-aware training strategies have been proposed, these approaches often fail to generalize on out-of-distribution data and can come at the cost of reduced overall performance~\cite{NatureYang2024}.
Currently, one of the most reliable approaches to improving both performance and fairness involves scaling the size and diversity of training datasets to better represent all patient demographic subgroups~\cite{chexclusion}.
However, privacy regulations and institutional barriers limit the feasibility of aggregating large, multi-site datasets, and such efforts have seen limited success to date. 
These challenges highlight the need for novel methods that can enhance model performance and fairness simultaneously, while operating within real-world constraints on data availability and sharing.

\begin{figure}[!htb]
  \centering
  \includegraphics[width=\textwidth]{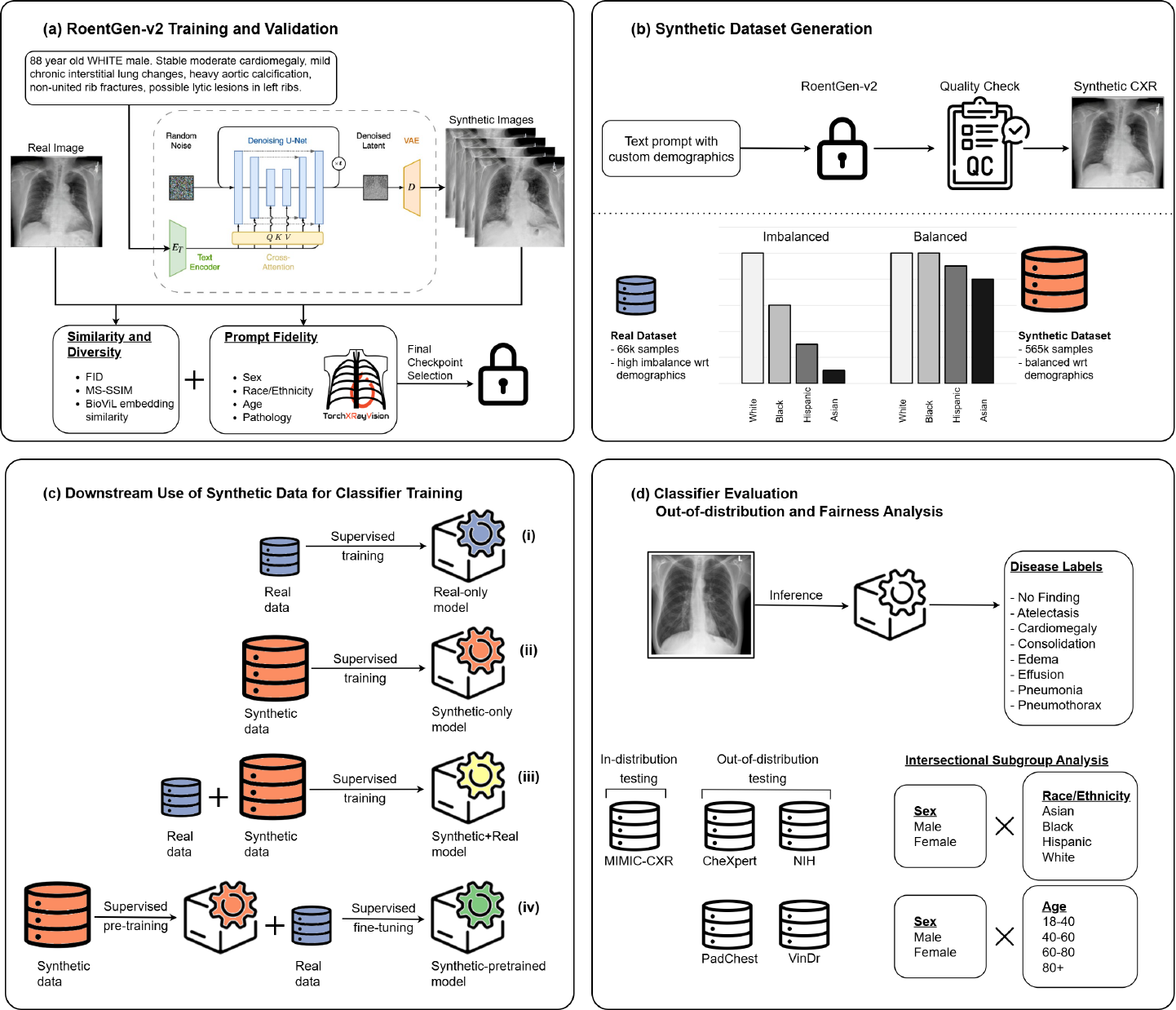}
  \caption{Overview of the experimental approach. 
  (a) \modelname is created by fine-tuning Stable Diffusion v2.1 on real chest radiographs (CXRs), using radiology report text enriched with demographic information. 
  (b) Top: A synthetic CXR dataset is generated by prompting the trained \modelname model with customized prompts, followed by quality control. Bottom: Comparison between the real training dataset and the generated synthetic dataset.
  (c) Classification models are trained using four approaches: 
  (i) Using only real CXRs;
  (ii) Using only synthetic CXRs generated by \modelname;
  (iii) Augmenting real CXRs with synthetic CXRs;
  (iv) Two-stage training, where we first perform supervised pretraining with synthetic CXRs, followed by fine-tuning on real CXRs.
  (d) The resulting classifiers are evaluated on multi-label classification tasks using both in-distribution and four out-of-distribution datasets, followed by intersectional subgroup analysis across sex, age, and race/ethnicity groups.}
  \label{fig1}
\end{figure}

Recent advances in generative diffusion models have significantly expanded the capabilities of deep learning in medical imaging~\cite{MllerFranzes2023,Khader2023}. 
Text-to-image models such as RoentGen~\cite{RoentgenBluethgen2024} and Cheff~\cite{cheff} have shown strong performance in synthesizing realistic CXRs with fine-grained control over disease-specific visual features. 
However, prior works leveraging diffusion-based synthetic data for CXR classification have several limitations. 
Thus far, conditioning has been limited to textual prompts describing radiographic findings, with little or no ability to control for demographic attributes such as age, sex, or race/ethnicity. 
This is a critical gap, given that computer vision models have been shown to detect and rely on demographic features, both explicitly and implicitly, when making predictions~\cite{Gichoya2022}. 
Consequently, the lack of demographic conditioning in earlier synthetic datasets limits their utility for studying or mitigating fairness-related concerns.
Furthermore, while synthetic data augmentation has been shown to improve model performance both in-distribution and on external test sets~\cite{LancetKhosravi2024,DeepmindKtena2024}, the optimal strategies for integrating synthetic data into model development pipelines remain unclear. 
Open questions persist around building data-centric pipelines for ensuring fidelity and relevance of synthetic samples for specific downstream clinical tasks, and how to best combine synthetic and real data.

In this work, we hypothesized that diffusion models with text and demographic conditioning can be used to generate large-scale synthetic CXR datasets with balanced representation across demographic subgroups, thereby enabling more equitable and generalizable training for downstream disease classification tasks. 
We further investigated how synthetic data should best be integrated into disease classifier development pipelines, specifically whether supervised pretraining on synthetic images is more effective than using synthetic data for augmentation during task-specific training.
Pretraining on diverse synthetic data has the potential to support the learning of robust, demographically invariant representations, thereby improving both performance and fairness across populations.

Our study makes four key contributions (Figure~\ref{fig1}).
First, we introduce \modelname, a state-of-the-art text-to-image diffusion model for CXRs, capable of high-fidelity image generation, with conditioning not only on radiographic findings but also on structured patient metadata, including sex, age, and race/ethnicity. 
Second, we used \modelname to create a large, demographically balanced synthetic dataset comprising over $565,000$ images. To ensure instruction-following of conditioning criteria, we integrated a quality control module into the generation pipeline that automatically filtered out inconsistent samples. 
Third, we conducted a comprehensive comparison of how synthetic data can be used to train disease classifiers, evaluating four strategies:
(i) Using only real CXRs;
(ii) Using only synthetic CXRs generated by \modelname;
(iii) Augmenting real CXRs with synthetic CXRs as is common in the literature;
(iv) A two-stage data-centric training, where we first perform supervised pretraining with synthetic CXRs, followed by fine-tuning on real CXRs.
Fourth, through systematic evaluation across five international datasets, we demonstrate that our two-stage supervised pretraining with synthetic data significantly improves both classification performance and fairness metrics across demographic subgroups. This approach consistently outperformed the other three training techniques.

Collectively, our findings show that high-quality, demographically controllable synthetic data are a powerful tool for improving performance and fairness in medical imaging deep learning, particularly in settings where real-world data are imbalanced, limited, or challenging to access. We open source \modelname weights, our synthetic dataset, and corresponding code to facilitate future research.

\begin{figure}[htbp]
    \centering
    \includegraphics[width=0.9\textwidth]{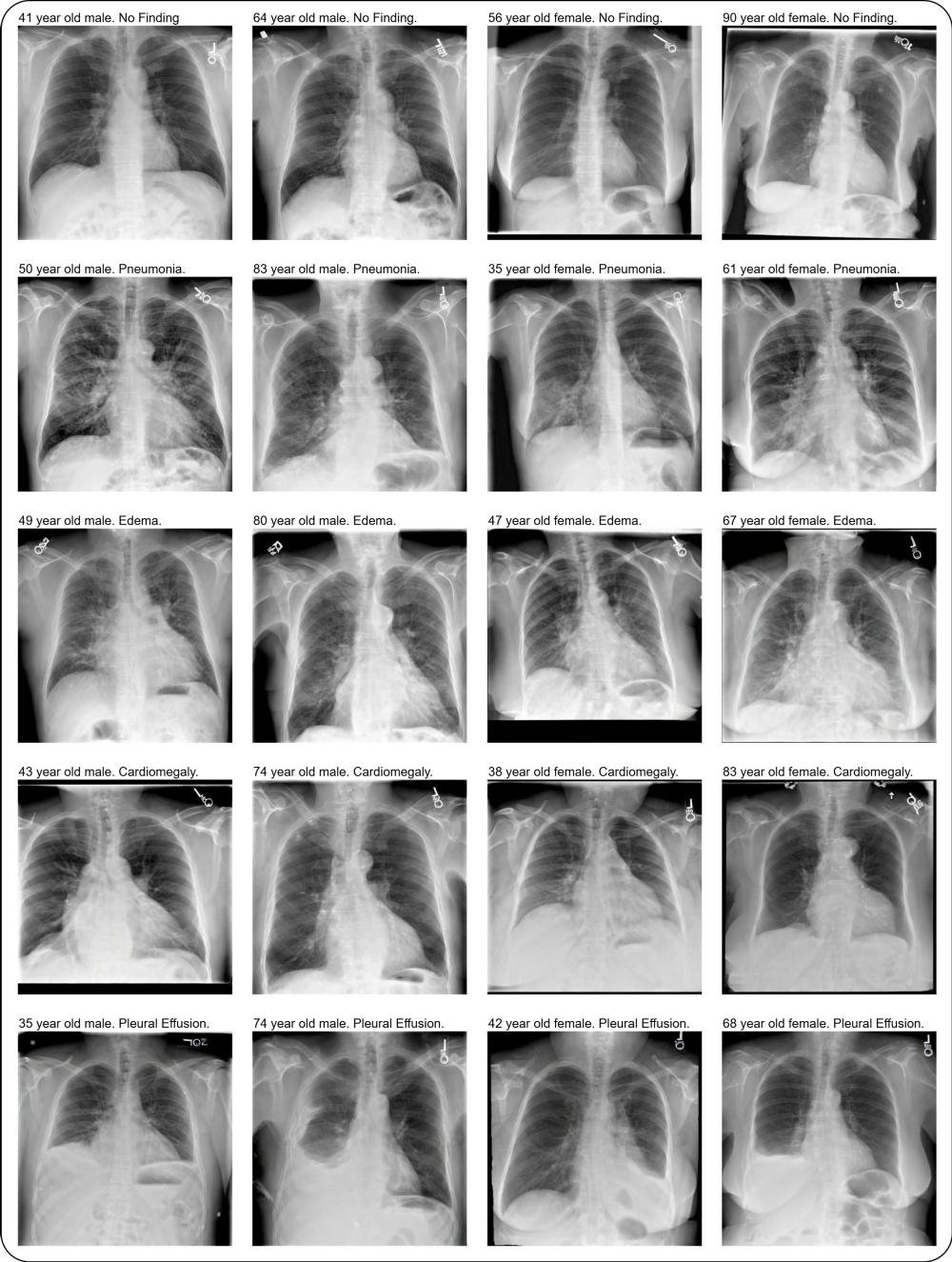}
    \caption{Example synthetic chest radiographs (CXRs) conditioned on radiographic findings and demographic information. Each row displays a distinct radiographic finding: No Finding, Pneumonia, Edema, Cardiomegaly and Pleural Effusion. Columns demonstrate variation across sex and age groups. The generated images maintain high fidelity and accurately follow the conditioning text prompts.}
    \label{fig_visuals}
\end{figure}


\section{Results}\label{sec2}

\subsection{Generative Model with Demographic Conditioning}
In this work, we present \modelname, which can condition image generation not only on radiographic findings, but also on patient demographic attributes (sex, age, race/ethnicity). Compared to the original RoentGen~\cite{RoentgenBluethgen2024} model, we use a $\sim$2$\times$ larger training dataset and use an improved baseline Stable Diffusion model (version 2.1 compared to version 1.4) to allow demographic-based metadata conditioning. 
We create a $16\times$ larger and demographically-balanced synthetic dataset ($565,154$ studies) to further rigorously evaluate data-centric approaches for best using synthetic data for training downstream classifiers with high accuracy, generalizability, and fairness. 
\modelname was trained on a curated subset of the MIMIC-CXR~\cite{mimic-Johnson2019} dataset for which images, radiology reports and patient metadata were available. 
We used the impression section, a summary section of a radiology report focused on the relevant findings and their interpretation.
We incorporated demographic information in the text prompt following the structure \texttt{<AGE> year old <RACE> <SEX>. <IMPRESSION>}.
\modelname was trained on a total of $68k$ CXR-report pairs, compared to RoentGen~\cite{RoentgenBluethgen2024} which used a training set of $35k$ CXR-report pairs.

To select the best generative model checkpoint we computed several metrics to measure the quality of synthetic images across three distinct axes.
(a) Alignment with the provided findings text and demographics from the prompt was evaluated by using pretrained classification models trained on real images (torch XRV \cite{Cohen2022xrv}) to predict labels for the synthetic images.
(b) Similarity of generated images to real images was evaluated using Fr\'{e}chet Inception Distance (FID).
(c) Diversity among synthetic images generated from the same text prompt was evaluated using the image-level multi-scale structural similarity index (MS-SSIM) and the cosine similarity of BioViL~\cite{BiovilBoecking2022} image embeddings. 
Supplementary Table~\ref{supptab1} summarizes the synthetic image quality metrics at each model checkpoint. 
Based on these criteria, we selected $10k$ training steps (equivalent to 28 epochs) as the final \modelname checkpoint, which required around 14 A100 GPU-hours for training. 
Detailed numerical data for all these criteria for checkpoint selection are provided in Supplementary Results~1. 

Briefly, for our chosen checkpoint, \modelname had an average disease area under the receiver operating characteristic curve (AUROC) of $0.81$, compared to $0.88$ for the real data baseline with XRV classifiers. 
As comparison, the original RoentGen~\cite{RoentgenBluethgen2024} model achieved an average disease AUROC of $0.82$. 
For demographic attributes, sex and race accuracy for synthetic images were $100\%$ and $98\%$, respectively (with original XRV sex and race classifier accuracy for the real data being $97\%$ and $95\%$, respectively). 
For the task of age prediction, we obtain comparable root mean square error ($8.9$ years) as the original XRV age regressor errors ($7.1$ years).
\modelname had an FID score of 76.8, which was considerably lower than the original RoentGen~\cite{RoentgenBluethgen2024} model score of 96.1. 
Lastly, when using the same prompts for image generation, we observe a low MS-SSIM of 0.37 and moderate BioViL similarity of 0.66, indicating diversity in the generated images.

Training time for the maximum number of $60k$ optimization steps (equivalent to 172 epochs) was 80 A100 GPU-hours. 

\subsection{Synthetic Dataset}

\begin{figure}[!htb]
\centering
\includegraphics[width=\textwidth]{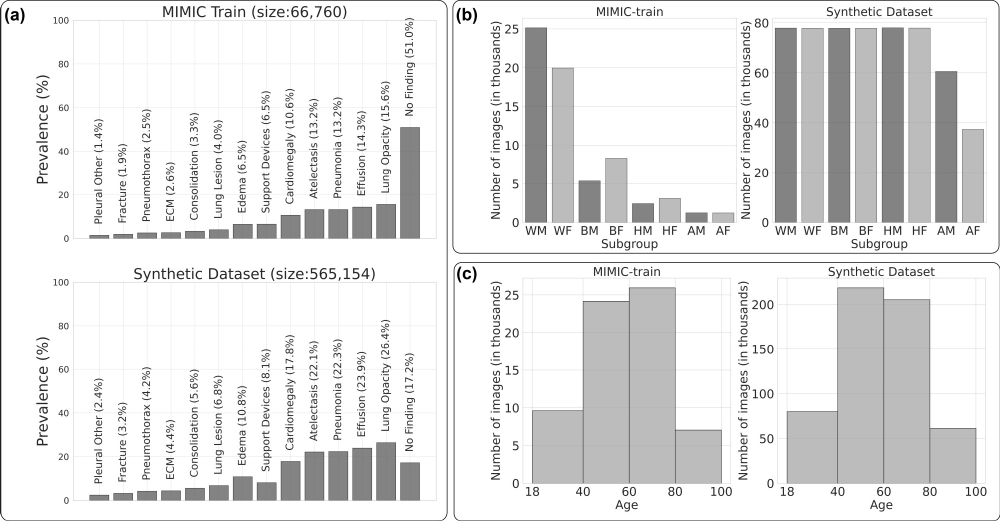}
\caption{Overview of label statistics for the synthetic dataset compared with the training dataset of the generative model. Panel (a) shows the prevalence for each of 13 diseases plus the `No Finding' label. ECM stands for Enlarged Cardiomediastinum. Panel (b) shows the subgroup composition in terms of sex (M:male, F:female) and race/ethnicity (A:Asian, B:Black, H:Hispanic, W:White). Panel (c) shows the binned age distributions.}\label{fig2}
\end{figure}

The best \modelname checkpoint was used to generate a large and demographically balanced synthetic dataset of $565,154$ samples.
For each radiology report impression section in the MIMIC-CXR train set, we created multiple custom text prompts, with all possible combinations of sex (male, female) and race/ethnicity (Asian, Black, Hispanic, White) categories.
For each new prompt variation, the age was randomly selected from a $\pm5$~year interval centered on the original age value. 

We implemented a quality check step in the inference pipeline to only accept an image if the prompt matched the output of pretrained sex, race and age classifiers~\cite{Cohen2022xrv}. Each failed prompt was re-generated a maximum of three times, then discarded.
We started with a total of $623,712$ prompts and ended with a quality controlled dataset of $565,154$ synthetic images.
A total of $58,558$ (9.4\%) prompts failed quality check on one or more demographic attributes more than three times and were discarded.
Of the 9.4\% of prompts that failed to converge, 9.2\% failed due to the race criterion, 0.2\% due to the age criterion; none failed as a result of the sex criterion.
This mode of failure is likely driven by the lack of direct imaging correlates of race as well as pronounced class imbalance in race/ethnicity within the generative model’s training data. 
Among the quality-controlled synthetic images, $467,914$ (83\%) corresponded to text reports with one or more positive disease findings, and $97,240$ (17\%) displayed normal CXRs.
Figure~\ref{fig2} shows a side-by-side comparison between the disease label distribution and demographics of the MIMIC-CXR train set and the synthetic dataset.
Overall, based on our available inference compute budget and the data diversity of the training set, we successfully generated the following numbers of correctly classified synthetic CXRs: $155,728$ for White demographic ($3.5\times$ increase from the original $45,102$ real White CXRs),
$155,653$ for Black demographic ($11\times$ increase from the original $13,685$ real Black CXRs),
$155,870$ for Hispanic demographic ($28\times$ increase from the original $5,521$ real Hispanic CXRs),
and $97,903$ for Asian demographic ($40\times$ increase from the original $2,452$ real Asian CXRs). 

Total inference time, including the quality check step, was 360 A100 40GB GPU-hours for generating the final $565,154$ synthetic images.

\subsection{Disease Classifiers}
We extend prior work to train CXR disease classifiers using synthetic data that includes disease and demographic information. 
Moreover, unlike prior work that naively combined synthetic and real datasets for training classifiers, we explore and identify improved mechanisms for training with synthetic data to optimize for overall classifier accuracy and fairness.
We focus on out-of-distribution (OOD) evaluation using posteroanterior (PA) view CXRs from four large external datasets: CheXpert~\cite{chexpert,chexpert-plus}, NIH~\cite{nih-Wang2017}, PadChest~\cite{padchest-Bustos2020} and VinDr~\cite{vindr-Nguyen2022}. 
We further conduct a detailed fairness analysis by assessing intersectional subgroups of sex \& race/ethnicity and sex \& age across multiple datasets.

\clearpage
\subsubsection{Synthetic Data Augmentation}

\begin{figure}[!htb]
\centering
\includegraphics[width=\textwidth]{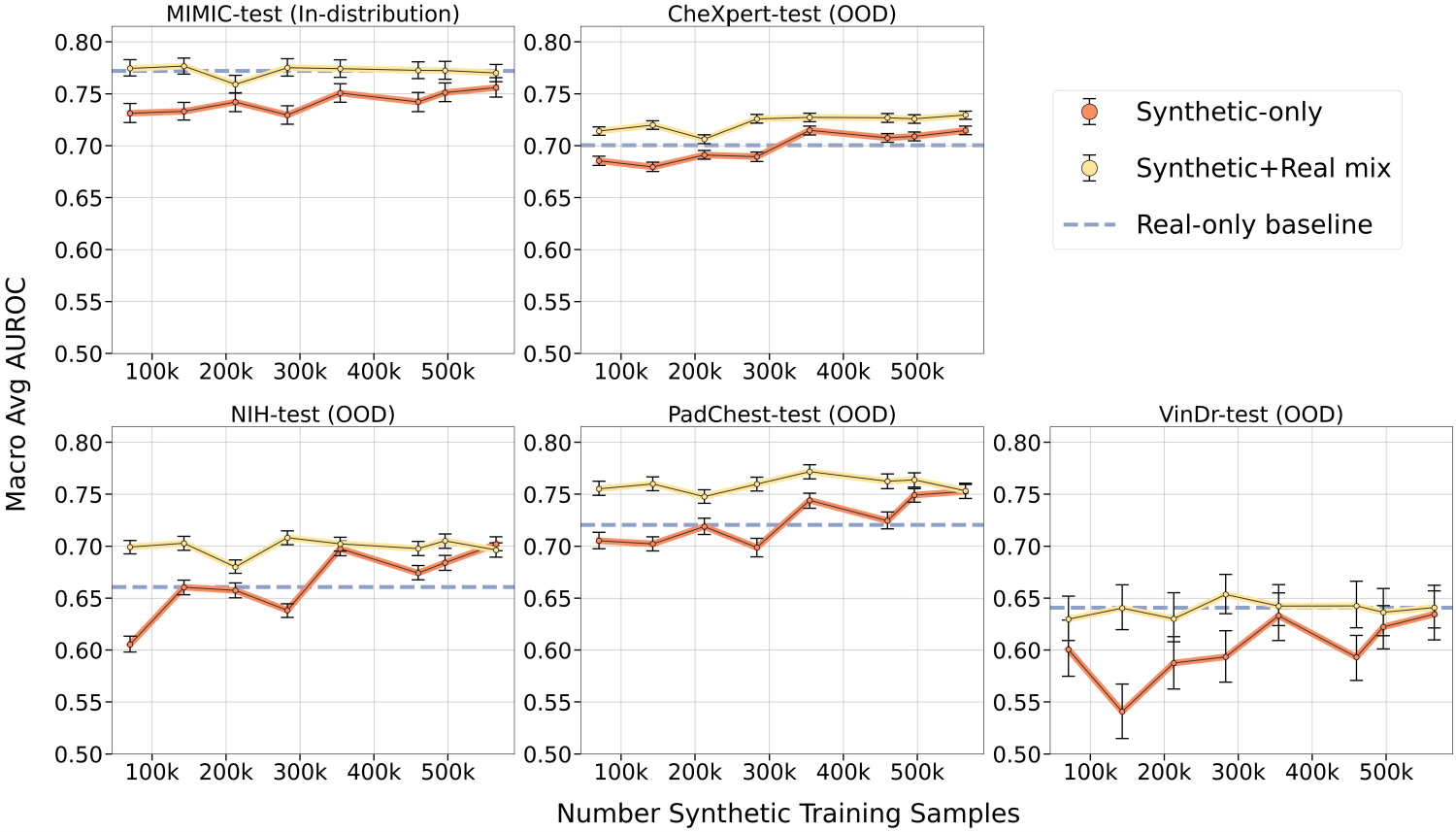}
\caption{Classification performance of models trained with increasing amounts of synthetic data, from the size of the real training dataset ($66k$ samples) up to $8\times$ the real data ($565k$ samples).
Synthetic-only models, shown in orange, are initialized from ImageNet pretrained weights and trained with varying amounts of synthetic images. 
Synthetic+Real mix models, shown in yellow, are initialized from ImageNet pretrained weights and trained using a combination of $66k$ real images plus increasing amounts of synthetic images.
Blue dashed line corresponds to the baseline model, initialized from ImageNet pretrained weights and trained with $66k$ real images only.
Y-axis shows average AUROC, error bars indicate 95\% confidence intervals; X-axis shows the number of synthetic images in the train set. 
Each model is evaluated in-distribution on MIMIC-CXR test split and out-of-distribution (OOD) on four external datasets.}\label{fig3}
\end{figure}

\paragraph{Training with a mix of real and synthetic data}
We trained downstream classifiers starting from ImageNet-pretrained weights and used a training set of real data from MIMIC-CXR combined with synthetic data from \modelname (termed the synthetic+real model).
We incrementally added synthetic data to the $66k$ real samples in the MIMIC-CXR training set, up to a maximum number of $565k$ synthetic images. 
The synthetic+real model with maximal data achieved an AUROC of 0.770 (CI: $0.762-0.778$) for the multi-label classification task on the in-distribution MIMIC-CXR test set.
This was on par with the real-only baseline which had an AUROC of 0.772 (CI: $0.765-0.781$), and the difference between the two was not statistically significant (p-value$=0.67$).

Training on synthetic+real data resulted in increased AUROC scores on out-of-distribution datasets compared to the real-only baseline: 
$4.3\%$ increase on CheXpert from 0.700 (CI: $0.696-0.705$) to 0.730 (CI: $0.725-0.733$), 
$5.3\%$ increase on NIH from 0.661 (CI: $0.654-0.668$) to 0.696 (CI: $0.690-0.703$), 
and $4.4\%$ increase on PadChest from 0.721 (CI: $0.713-0.729$) to 0.753 (CI: $0.746-0.761$), 
all for maximal data supplementation; differences were statistically significant (p-value$<0.05$).
On the VinDr test set, the synthetic+real model achieved an AUROC of 0.641 (CI: $0.621-0.662$), on par with the real-only baseline AUROC of 0.641 (CI: $0.617-0.663$), with no significant difference between the two (p-value$=0.99$).
Supplementing real training data with synthetic training data improved performance on most out-of-distribution datasets. 
This effect plateaued at $300k$ synthetic samples, or about $5\times$ the number of real data samples.
Figure~\ref{fig3} summarizes these findings for all the intermediary data supplementation steps.

\paragraph{Training only with synthetic data}
When classifiers were trained using synthetic data alone, multi-label AUROC in-distribution remained marginally lower that of the real-only baseline.
The synthetic-only model trained on $565k$ images obtained an AUROC of 0.756 (CI: $0.747-0.766$) on MIMIC-test, slightly below the real-only baseline of 0.772 (CI: $0.765-0.781$); the difference was significant with p-value$<0.05$. 
However, synthetic-only models trained with $300k$ or more images were able to match or surpass the performance of the real-only baseline on out-of-distribution evaluations. 
The synthetic-only model trained on $565k$ images achieved the following multi-label AUROCs on external datasets:
on CheXpert, 0.715 (CI: $0.711-0.719$) compared to real-only baseline of 0.700 (CI: $0.696-0.705$); 
on NIH, 0.702 (CI: $0.696-0.709$) compared to real-only baseline of 0.661 (CI: $0.654-0.668$);
on PadChest, 0.753 (CI: $0.746-0.759$) compared to real-only baseline of 0.721 (CI: $0.713-0.729$); 
on VinDr, 0.635 (CI: $0.610-0.657$) compared to real-only baseline of 0.641 (CI: $0.617-0.663$).
All differences were significant with p-value$<0.05$, except for the VinDr test set where p-value=$0.70$.

When directly comparing the synthetic-only model (trained on $565k$ generated images) with the synthetic+real mix model (trained on $565k$ generated images and $66k$ real images), there was no significant difference in AUROC on NIH (p-value$=0.19$), PadChest (p-value$=0.93$) and VinDr (p-value=$0.65$). 
On MIMIC-test, the synthetic+real model achieved slightly higher AUROC, 0.770 (CI: $0.762-0.778$) compared to the synthetic-only model, 0.756 (CI: $0.747-0.766$), with an associated p-value$<0.05$.
On CheXpert-test, the synthetic+real model achieved 0.730 (CI: $0.725-0.733$), slightly higher than the synthetic-only model score of 0.715 (CI: $0.711-0.719$), with a corresponding p-value$<0.05$.
When using fewer generated images in the train set, differences between the synthetic+real and synthetic-only models were more pronounced, as illustrated in Figure~\ref{fig3}. 
Overall, we unsurprisingly find that training on synthetic data can always benefit from the addition of real data.

\subsubsection{Synthetic pretraining}

\begin{figure}[!htb]
\centering
\includegraphics[width=\textwidth]{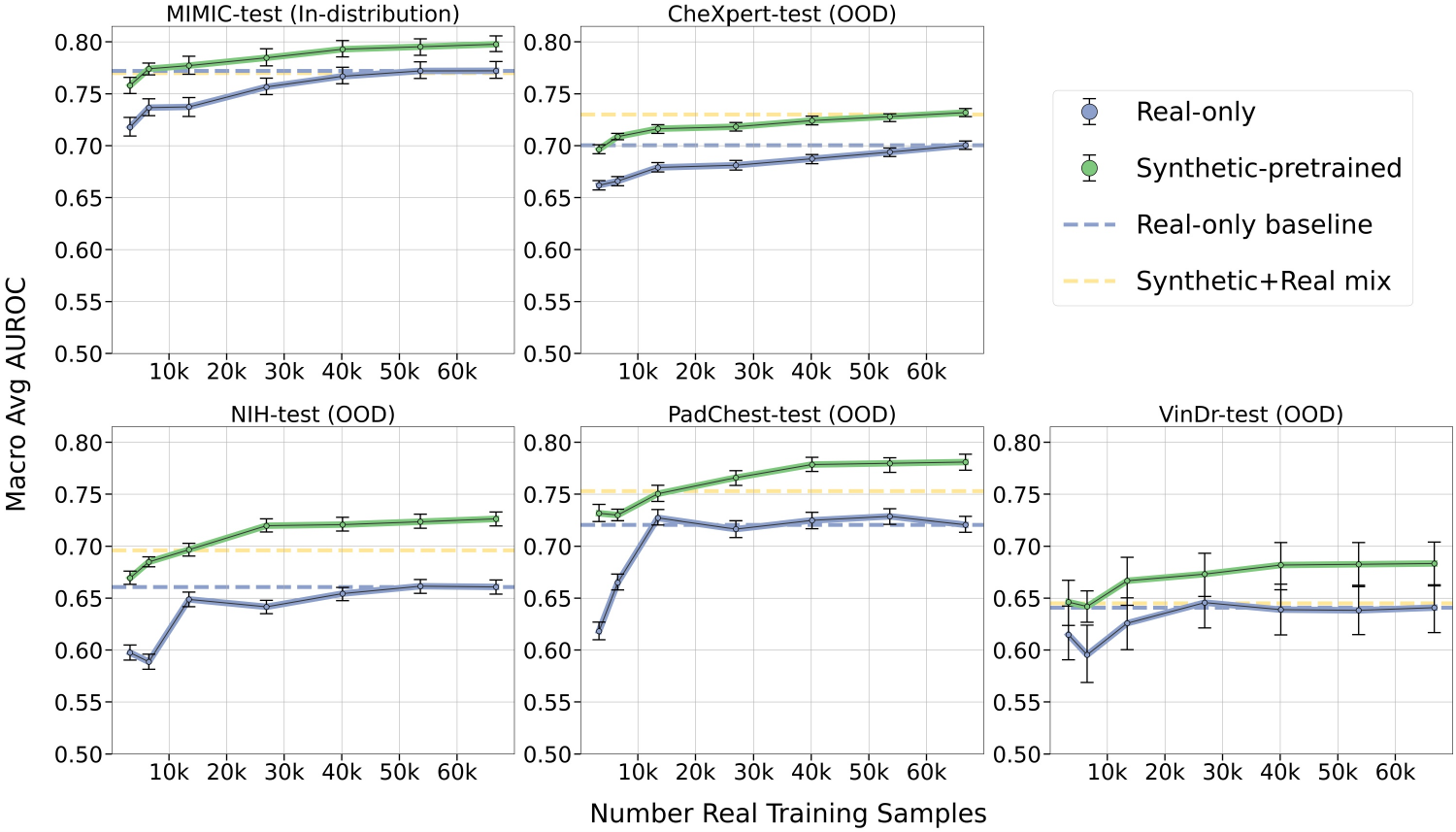}
\caption{
Classification performance of models trained with varying amounts of real data, from the maximally available $66k$ samples down to data-scarce regimes with as little as $3k$ samples.
Real-only models, shown in blue, are initialized from ImageNet pretrained weights and trained with varying amounts of real images.
The blue dashed line corresponds to the baseline model trained with all $66k$ real images.
Synthetic-pretrained models, shown in green, are first pretrained from scratch with $565k$ synthetic images, then fine-tuned with increasing amounts of real images.
Yellow dashed line corresponds to the Synthetic+Real mix model, initialized from ImageNet pretrained weights and trained with all available data ($66k$ real images and $565k$ synthetic images).
Y-axis shows average AUROC, error bars indicate 95\% confidence intervals; x-axis shows the number of real images in the train set. 
Each model is evaluated in-distribution on MIMIC-test and out-of-distribution (OOD) on four external datasets.}\label{fig4}
\end{figure}

It is common practice in the computer vision literature to initialize models with weights pretrained on natural image datasets such as ImageNet. 
We propose an analogous supervised pretraining strategy that performs pretraining with synthetic CXR data instead of natural images, followed by supervised fine-tuning on MIMIC-CXR train set (real data only).
Figure~\ref{fig4} illustrates the benefits of synthetic pretraining when fine-tuning classifiers on varying amounts of real data.
Synthetic pretraining led to consistent AUROC improvements over the real-only baseline (initialized from ImageNet weights) both in- and out-of-distribution.
When fine-tuning the synthetic-pretrained model using the maximum available real data ($66k$ images), we observed the following performance gains:
on MIMIC-CXR a 3.3\% increase from 0.772 (CI: $0.765-0.781$) to 0.798 (CI: $0.791-0.806$), 
on CheXpert a 4.5\% increase from 0.700 (CI: $0.696-0.705$) to 0.732 (CI: $0.728-0.736$), 
on NIH a 9.8\% increase from 0.661 (CI: $0.654-0.668$) to 0.726, (CI: $0.720-0.733$), 
on PadChest a 8.3\% increase from 0.721 (CI: $0.713-0.729$) to 0.781 (CI: $0.773-0.789$),
and on VinDr a 6.5\% increase from 0.641 (CI: $0.617-0.663$) to 0.683 (CI: $0.662-0.704$). 
All differences were statistically significant with p-value$<0.05$.

Moreover, when using synthetic pretraining, by fine-tuning on less than $10k$ real samples, we match or exceed the AUROC of the real-only baseline trained on the full $66k$ real samples (illustrated by the dashed blue line in Figure~\ref{fig4}). 
Also, when using synthetic pretraining and fine-tuning on $30k$ or more real samples, we exceed the AUROC of the synthetic+real mix model trained on the combination of all available data (illustrated by the dashed yellow line in Figure~\ref{fig4}).
This improvement was statistically significant (p-value$<0.05$) on MIMIC, NIH, PadChest, and VinDr, but not on CheXpert test dataset.
Overall, pretraining with synthetic data can improve performance over ImageNet pretraining or can achieve the same performance with dramatic reductions in the need for real training data.
Furthermore, two-stage training that combines supervised pretraining with synthetic data followed by fine-tuning with real data outperforms naively mixing synthetic and real data together, as is the norm in prior studies. 

\begin{figure}[!htb]
\centering
\includegraphics[width=\textwidth]{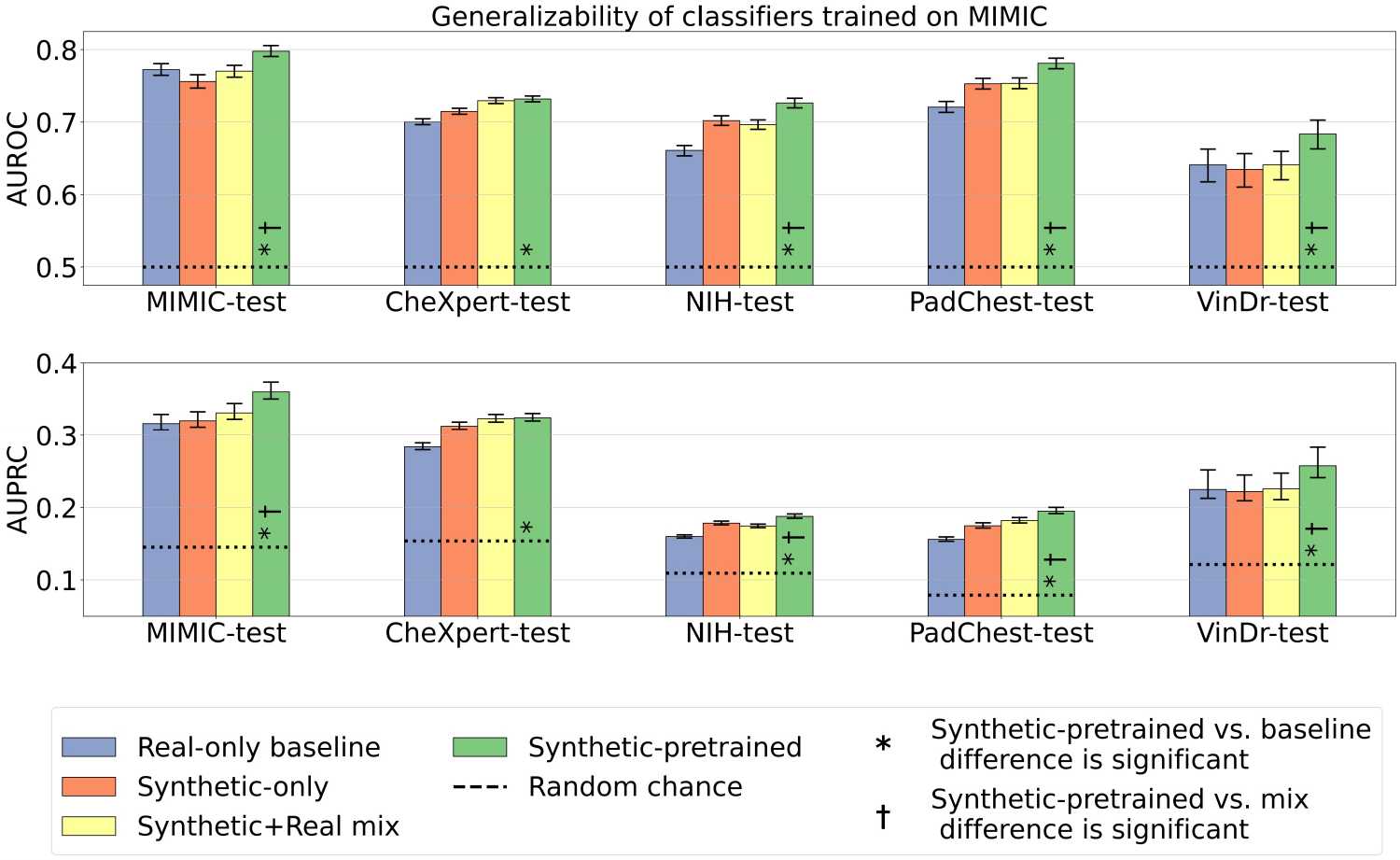}
\caption{Classification performance of models trained on all available real and/or synthetic data according to four strategies: 
(i) Real-only (baseline) model trained on $66k$ real CXRs, 
(ii) Synthetic-only model trained on $565k$ synthetic CXRs, 
(iii) Synthetic+Real mix model trained on combined $66k$ real and $565k$ synthetic CXRs, and 
(iv) Synthetic-pretrained model, which underwent supervised pretraining on $565k$ synthetic CXRs followed by fine-tuning on $66k$ real CXRs.
In scenarios (i)--(iii) models were initialized using ImageNet pretrained weights; in scenario (iv) the model was trained from scratch. 
Y-axis shows macro-average multi-label AUROC (top panel), and AUPRC (bottom panel), with error bars indicating 95\% confidence intervals. 
Each model is evaluated in-distribution on MIMIC-test and out-of-distribution on four external datasets.}\label{fig5}
\end{figure}

Figure~\ref{fig5} shows the overall AUROC comparison between the best performing models from each training strategy, using the maximally available training data: 
(i) Real-only baseline model (trained on $66k$ real CXRs), 
(ii) Synthetic-only model (trained on $565k$ synthetic CXRs), 
(iii) Synthetic+Real mix model (trained on combined $66k$ real CXRs and $565k$ synthetic CXRs), and 
(iv) Synthetic-pretrained model (supervised pretraining on $565k$ synthetic CXRs, followed by fine-tuning on $66k$ real-only CXRs). 
In scenarios (i)--(iii) models were initialized using ImageNet pretrained weights; in scenario (iv) the model was trained from scratch. 

The synthetic pretraining strategy yielded the best multi-label AUROC performance across all datasets. 
In-distribution, on MIMIC-test, the synthetic-pretrained model achieved an AUROC of 0.798 (CI: $0.791-0.806$), above the real-only baseline of 0.772 (CI: $0.765-0.781$) and the synthetic+real mix model of 0.770 (CI: $0.762-0.778$).
Out-of-distribution, on CheXpert-test, the synthetic-pretrained model achieved an AUROC of 0.732 (CI: $0.728-0.736$), on par with the  synthetic+real mix model of 0.730 (CI: $0.725-0.733$), both above the real-only baseline of 0.700 (CI: $0.696-0.705$).
On NIH-test, the synthetic-pretrained model achieved an AUROC of 0.726, (CI: $0.720-0.733$), above the synthetic+real mix model of 0.696 (CI: $0.690-0.703$), and the real-only baseline of 0.661 (CI: $0.654-0.668$).
On PadChest-test, the synthetic-pretrained model achieved an AUROC of 0.781 (CI: $0.773-0.789$), above the synthetic+real mix model of 0.753 (CI: $0.746-0.761$), and the real-only baseline of 0.721 (CI: $0.713-0.729$).
Finally, on VinDr-test, the synthetic-pretrained model achieved an AUROC of 0.683 (CI: $0.662-0.704$), above the real-only baseline of 0.641 (CI: $0.617-0.663$) and the synthetic+real mix model of 0.641 (CI: $0.621-0.662$).
All pairwise AUROC differences between the synthetic-pretrained model and other models were statistically significant with p-value$<0.05$ under the DeLong test unless specified otherwise.
Supplementary Figure~\ref{fig:suppfig1} shows disease-specific AUROC metrics across all models and datasets (discussed in Supplementary Note 1).

The synthetic pretraining strategy also yielded the highest multi-label area under the precision recall curve (AUPRC) values. 
In-distribution, on MIMIC-test, the synthetic-pretrained model achieved an AUPRC of 0.360 (CI: $0.350-0.374$), above the synthetic+real mix model of 0.331 (CI: $0.322-0.344$) and the real-only baseline of 0.316 (CI: $0.307-0.329$).
Out-of-distribution, on CheXpert-test, the synthetic-pretrained model achieved an AUPRC of 0.324 (CI: $0.320-0.330$), on par with the  synthetic+real mix model of 0.323 (CI: $0.318-0.329$), both above the real-only baseline of 0.284 (CI: $0.280-0.289$).
On NIH-test, the synthetic-pretrained model achieved an AUPRC of 0.188, (CI: $0.185-0.191$), above the synthetic+real mix model of 0.174 (CI: $0.172-0.177$), and the real-only baseline of 0.160 (CI: $0.158-0.162$).
On PadChest-test, the synthetic-pretrained model achieved an AUPRC of 0.195 (CI: $0.191-0.200$), above the synthetic+real mix model of 0.182 (CI: $0.179-0.186$), and the real-only baseline of 0.156 (CI: $0.153-0.159$).
Finally, on VinDr-test, the synthetic-pretrained model achieved an AUPRC of 0.258 (CI: $0.241-0.283$), above the real-only baseline of 0.225 (CI: $0.212-0.252$) and the synthetic+real mix model of 0.226 (CI: $0.211-0.247$).
All pairwise AUPRC differences between the synthetic-pretrained model and other models were statistically significant with p-value$<0.05$ under the permutation test unless specified otherwise.
Supplementary Figure~\ref{fig:suppfig2} shows disease-specific AUPRC metrics across all models and datasets (discussed in Supplementary Note 1).

\subsubsection{Classifier Fairness}

Beyond boosting classifier performance and generalization with synthetic data, we also investigate how synthetic data impacts classifier fairness.
We measure fairness using the following metrics: 
(a) subgroup-level classification performance (multi-label AUROC),
(b) average performance gap between the best and worst performing subgroups as per~\cite{DeepmindKtena2024}, and 
(c) underdiagnosis gap between the best and worst performing subgroups as proposed by~\cite{NatureSeyyedKalantari2021,NatureYang2024}.
We analyzed intersectional subgroups, here defined as patients who belong to two subpopulations (e.g. a female patient of Hispanic ethnicity). 
This is motivated by prior work showing that intersectional subgroups often have compounded algorithmic biases~\cite{NatureSeyyedKalantari2021}. 
Based on the availability of metadata, we define subgroups both by intersection of sex \& race/ethnicity (e.g. Black female) and sex \& age (e.g. male aged 40-60 years old) for MIMIC-CXR and CheXpert. 
We define subgroups by intersection of sex \& age attributes for NIH and PadChest datasets. 
Since either sex and/or age data are missing for over 83\% of patients in VinDr-test, we do not perform subgroup analysis for this dataset.

\begin{figure}[!htbp]
\centering
\includegraphics[width=0.75\textwidth]{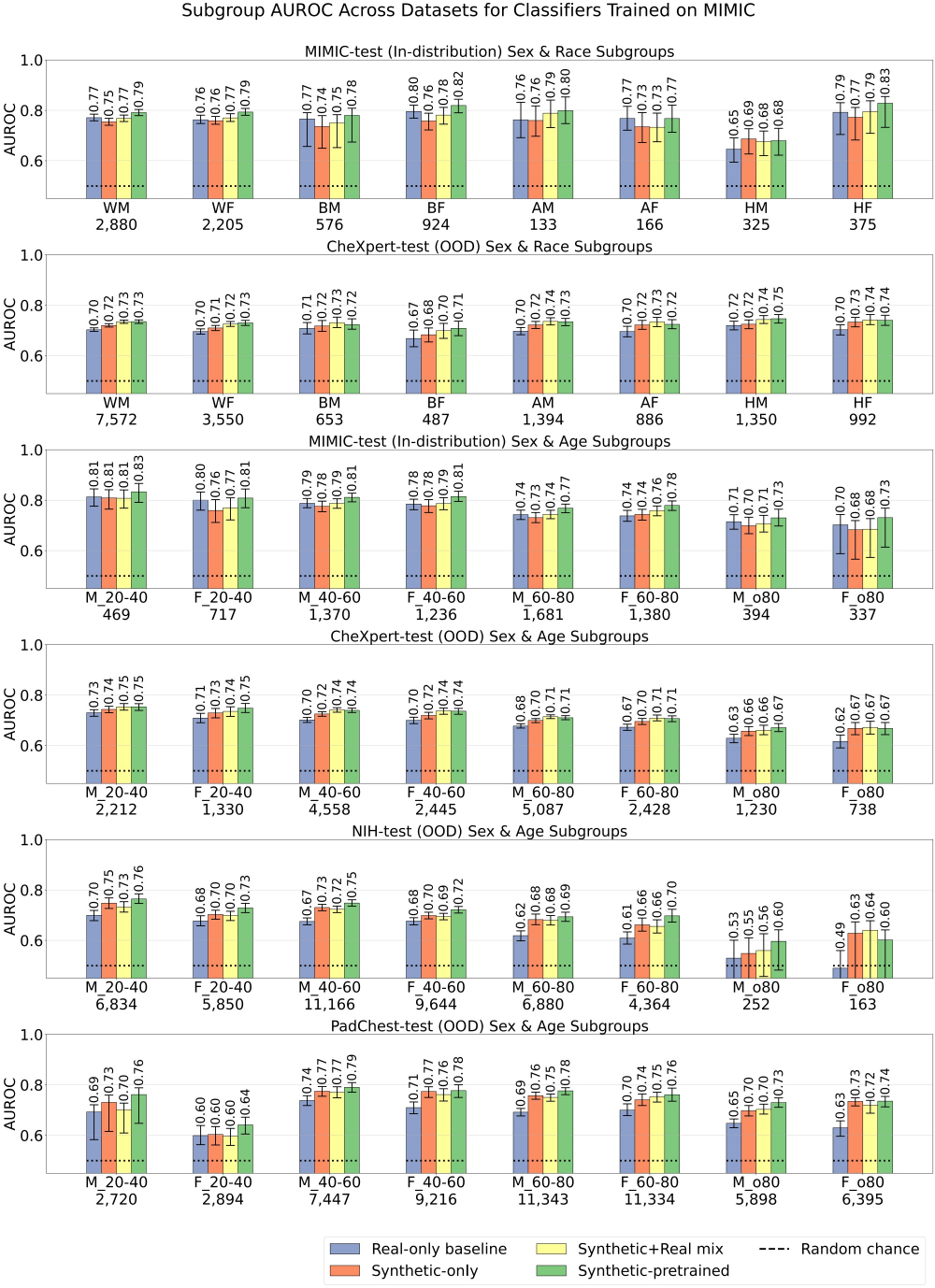}
\caption{Subgroup classification performance of models trained on all available real and/or synthetic data according to four strategies: 
(i) Real-only (baseline) model trained on $66k$ real CXRs, 
(ii) Synthetic-only model trained on $565k$ synthetic CXRs, 
(iii) Synthetic+Real mix model trained on combined $66k$ real and $565k$ synthetic CXRs, and 
(iv) Synthetic-pretrained model, which underwent supervised pretraining on $565k$ synthetic CXRs followed by fine-tuning on $66k$ real CXRs.
In scenarios (i)--(iii) models were initialized using ImageNet pretrained weights; in scenario (iv) the model was trained from scratch. 
Y-axis shows macro-average multi-label AUROC, with each panel focusing on one dataset. MIMIC-test is in-distribution, while CheXpert, NIH, and PadChest are out-of-distribution.
A random chance classifier would score 0.50 AUROC. 
Top two rows use subgroups defined by sex \& race/ethnicity (M:male, F:female, W:White, B:Black, H:Hispanic, A:Asian). 
Remaining rows use subgroups defined by sex \& age.
On the x-axis, below each subgroup's label is the patient count of the subgroup.}\label{figA3}
\end{figure}

\paragraph{(a) Subgroup Classification Performance}
For each model, we calculated the macro-average multi-label AUROC for all intersectional subgroups.
In Figure~\ref{figA3}, we depict the AUROCs of each subgroup for the best model in each training paradigm. 
We observe that synthetic pretraining reliably improves classification performance in all demographic subgroups across all datasets. 
For subgroups defined by sex \& race/ethnicity, the synthetic-pretrained model boosted performance over baseline in 7 out of 8 subgroups on MIMIC (there was no significant difference from baseline for the Asian female subgroup), and in all 8 out of 8 subgroups on CheXpert.
For subgroups defined by sex \& age, the synthetic-pretrained model boosted performance over baseline in all 8 out of 8 subgroups on MIMIC, CheXpert, NIH and PadChest.
Therefore, the proposed synthetic pretraining method maximally reduces fairness gaps by providing performance benefits irrespective of protected demographic attributes.

\paragraph{(b) Average Performance Gap}

\begin{figure}[!htb]
\centering
\includegraphics[width=0.635\textwidth]{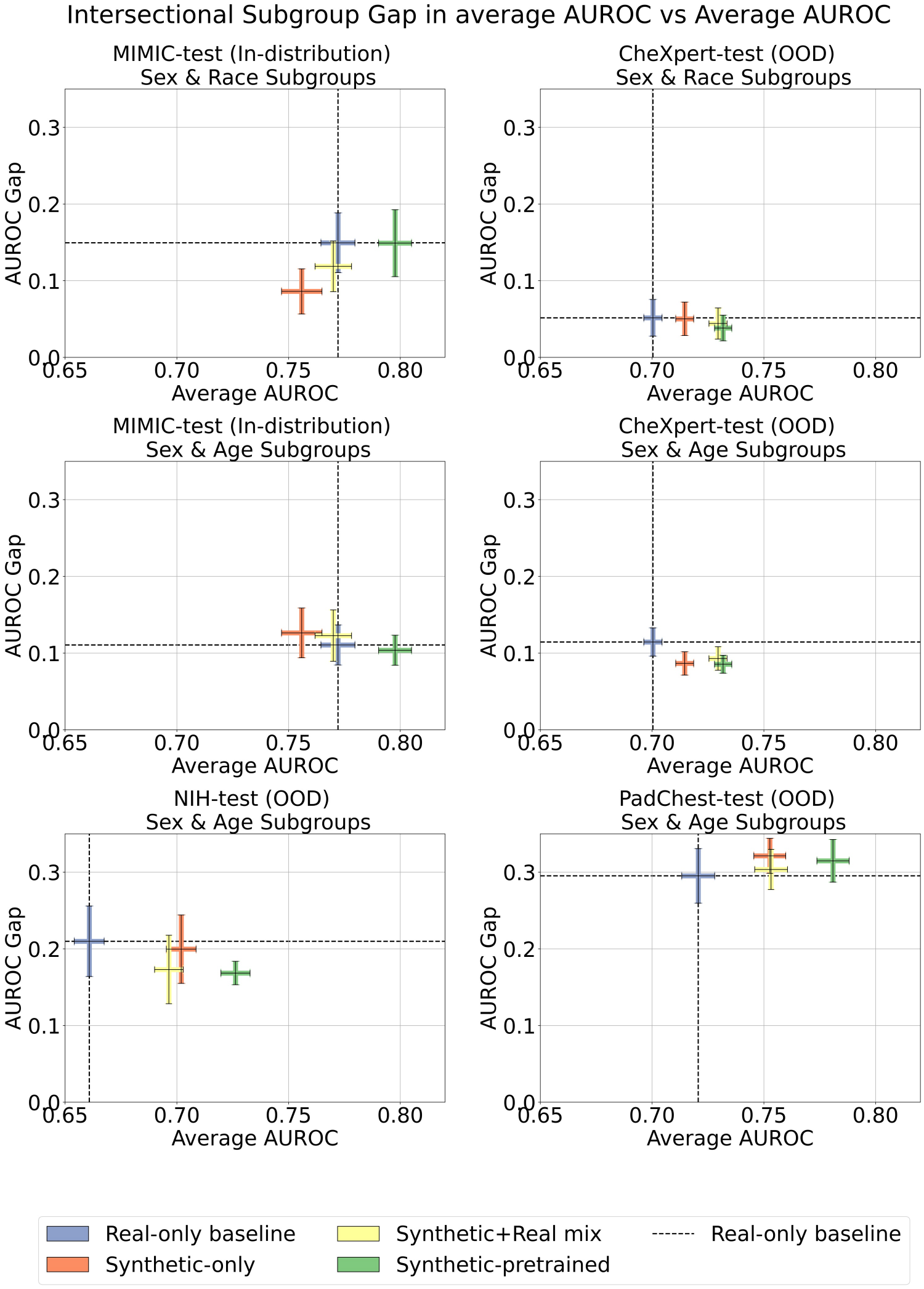}
\caption{Fairness gap in classifier AUROC across intersectional subgroups. 
Top row uses subgroups defined by sex \& race/ethnicity, middle and bottom rows use subgroups defined by sex \& age. 
Y-axis measures fairness gap between the best and worst performing subgroup for each model; lower values indicate better fairness. 
X-axis measures average model AUROC in the entire test population; higher values indicate better overall classification performance.
Models were trained on all available real and/or synthetic data according to four strategies: 
(i) Real-only (baseline) model trained on $66k$ real CXRs, 
(ii) Synthetic-only model trained on $565k$ synthetic CXRs, 
(iii) Synthetic+Real mix model trained on combined $66k$ real and $565k$ synthetic CXRs, and 
(iv) Synthetic-pretrained model, which underwent supervised pretraining on $565k$ synthetic CXRs followed by fine-tuning on $66k$ real CXRs.
In scenarios (i)--(iii) models were initialized using ImageNet pretrained weights; in scenario (iv) the model was trained from scratch.}\label{fig6}
\end{figure}

For each model, we calculated the average AUROC in all intersectional subgroups. 
We define the \emph{fairness gap} as the difference between the best and worst performing subgroups.
Similar to~\cite{DeepmindKtena2024}, we plot the fairness gap against average AUROC for all trained models, evaluated in-distribution and on out-of-distribution test sets (Figure~\ref{fig6}).
On average, for in-distribution evaluation on MIMIC, we observed little to no difference in the fairness gap between the baseline and synthetic-pretrained model, while multi-label AUROC improved by 3.3\%.
Out-of-distribution synthetic pretraining led to smaller fairness gaps by 16.0\% on average, while improving multi-label AUROC by an average 7.3\%.

For sex \& race/ethnicity subgroups, on MIMIC, the baseline fairness gap was 0.149 (CI: $0.110-0.188$) and the synthetic-pretrained fairness gap was 0.149 (CI: $0.105-0.193$).
On CheXpert, the baseline fairness gap was 0.051 (CI: $0.027-0.075$) and the synthetic-pretrained fairness gap was 0.038 (CI: $0.021-0.055$), a 25.5\% reduction.

For sex \& age subgroups, the synthetic-pretrained model achieved the smallest fairness gap on 3 out of 4 evaluation datasets.
On MIMIC, the baseline fairness gap was 0.110 (CI: $0.084-0.137$) and the synthetic-pretrained fairness gap was 0.104 (CI: $0.084-0.123$), a 5.5\% reduction.
On CheXpert, the baseline fairness gap was 0.114 (CI: $0.096-0.133$) and the synthetic-pretrained fairness gap was 0.085 (CI: $0.074-0.097$), a 25.4\% reduction.
On NIH, the baseline fairness gap was 0.210 (CI: $0.164-0.256$) and the synthetic-pretrained fairness gap was 0.168 (CI: $0.153-0.184$), a 20.0\% reduction.
Finally, on PadChest, the baseline fairness gap was 0.295 (CI: $0.259-0.331$) and the synthetic-pretrained fairness gap was 0.315 (CI: $0.287-0.342$), a 6.8\% increase.

\paragraph{(c) Underdiagnosis Gap}

\begin{figure}[!htb]
\centering
\includegraphics[width=0.635\textwidth]{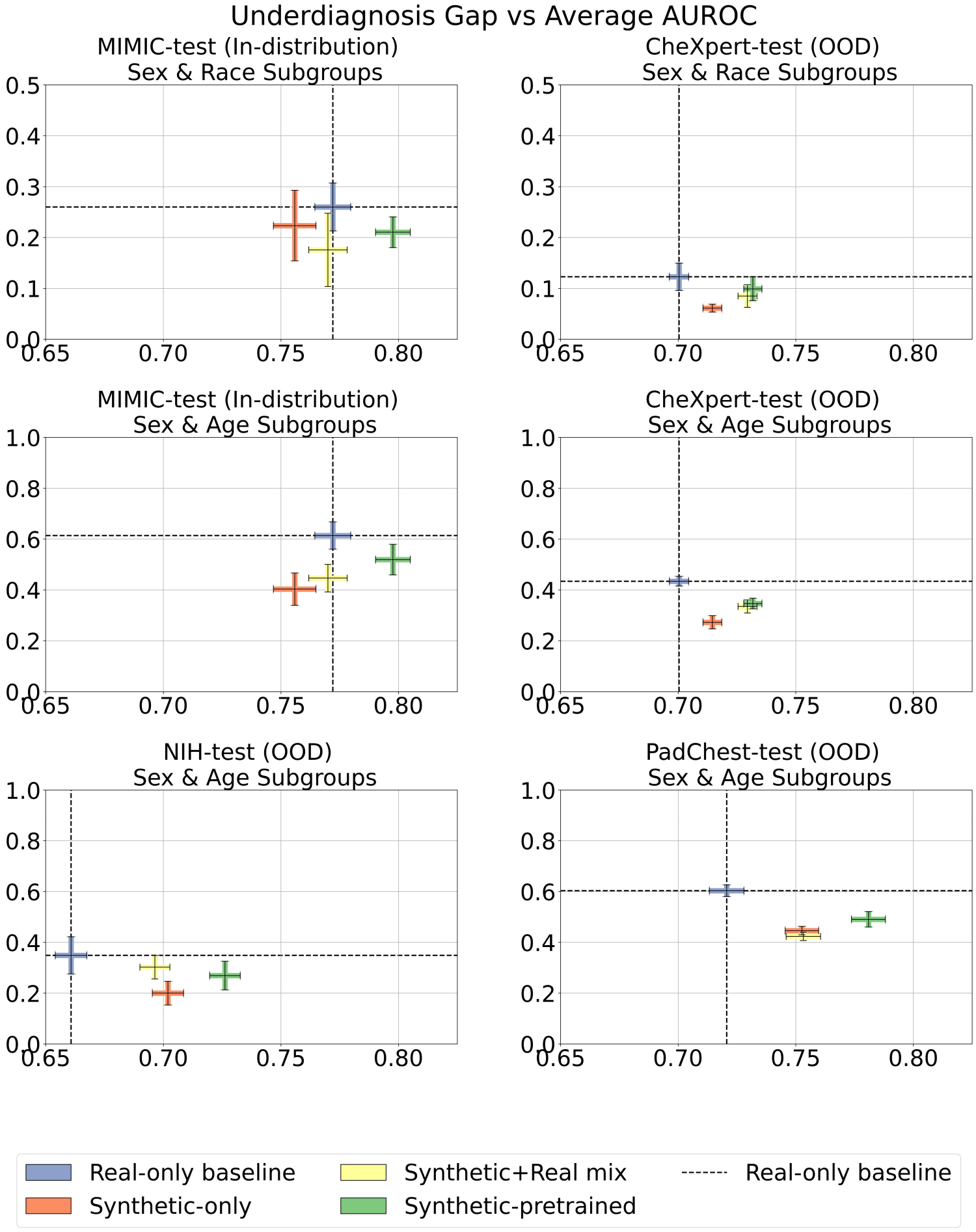}
\caption{Underdiagnosis gap in classifiers across intersectional subgroups. 
Top row uses subgroups defined by sex \& race/ethnicity, middle and bottom rows use subgroups defined by sex \& age. 
Y-axis measures underdiagnosis (False Positive Rate of `No Finding' label) gap between the best and worst performing subgroup for each model; lower values indicate better fairness. 
X-axis measures average model AUROC in the entire test population; higher values indicate better overall classification performance.
Models were trained on all available real and/or synthetic data according to four strategies: 
(i) Real-only (baseline) model trained on $66k$ real CXRs, 
(ii) Synthetic-only model trained on $565k$ synthetic CXRs, 
(iii) Synthetic+Real mix model trained on combined $66k$ real and $565k$ synthetic CXRs, and 
(iv) Synthetic-pretrained model, which underwent supervised pretraining on $565k$ synthetic CXRs followed by fine-tuning on $66k$ real CXRs.
In scenarios (i)--(iii) models were initialized using ImageNet pretrained weights; in scenario (iv) the model was trained from scratch.}\label{fig8}
\end{figure}

Multiple studies consider underdiagnosis as a primary criteria for fairness, due to its potentially harmful impact on patient outcomes, such as not receiving a treatment or receiving one with a delay. 
Following~\cite{NatureSeyyedKalantari2021,NatureYang2024}, we define the \emph{underdiagnosis rate} as the false-positive rate (FPR) of the binarized model prediction for the `No Finding' label, indicating a missed diagnosis. 
We compute this underdiagnosis rate at the levels of intersectional subgroups.
The corresponding \emph{underdiagnosis gap} is the difference between subgroups with the highest and lowest FPRs for each model.
Figure~\ref{fig8} plots the underdiagnosis gap against average model AUROC for the best model in each training paradigm.
The use of synthetic data, either as augmentation strategy or pretraining strategy reduced the underdiagnosis gap across all datasets compared to the real-only baseline.
On average across datasets, synthetic pretraining led to a 19.3\% decrease in the underdiagnosis gap, while increasing multi-label AUROC by 6.5\%.

For sex \& race/ethnicity subgroups on MIMIC, the baseline underdiagnosis gap was 0.260 (CI: $0.213-0.307$) and the synthetic-pretrained underdiagnosis gap was 0.210 (CI: $0.180-0.241$), a 19.2\% reduction.
On CheXpert, the baseline underdiagnosis gap was 0.123 (CI: $0.096-0.150$) and the synthetic-pretrained underdiagnosis gap was 0.099 (CI: $0.076-0.122$), a 19.5\% reduction.

For sex \& age subgroups, the synthetic-pretrained model achieved smaller underdiagnosis gaps on 4 out of 4 evaluation datasets.
The smallest underdiagnosis gaps were displayed by the synthetic-only models, followed by the synthetic+real mix models.
On MIMIC, the baseline underdiagnosis gap was 0.614 (CI: $0.560-0.667$) and the synthetic-pretrained underdiagnosis gap was 0.519 (CI: $0.459-0.579$), a 15.5\% reduction.
On CheXpert, the baseline underdiagnosis gap was 0.434 (CI: $0.415-0.453$) and the synthetic-pretrained underdiagnosis gap was 0.346 (CI: $0.325-0.367$), a 20.3\% reduction.
On NIH, the baseline underdiagnosis gap was 0.349 (CI: $0.275-0.422$) and the synthetic-pretrained underdiagnosis gap was 0.269 (CI: $0.213-0.325$), a 22.9\% reduction.
On PadChest, the baseline underdiagnosis gap was 0.603 (CI: $0.580-0.626$) and the synthetic-pretrained underdiagnosis gap was 0.490 (CI: $0.460-0.521$), a 18.7\% reduction.


\section{Discussion}\label{sec3}
Our study conducted a detailed investigation into how classifier performance and fairness generalize when leveraging a large, demographically balanced synthetic dataset. 
Specifically, we demonstrated that text-conditioned latent diffusion models can generate synthetic chest radiographs with fine-grained control over both radiographic findings and key demographic attributes, including sex, age and race/ethnicity. 
To ensure fidelity, we incorporated rigorous quality control measures into the inference pipeline, filtering out low-quality or misaligned samples to retain only those that accurately matched the conditioning prompts.

We trained downstream classifiers using four distinct strategies:
(i) Using only real CXRs; 
(ii) Using only synthetic CXRs generated
by RoentGen-v2 ; 
(iii) Augmenting real CXRs with synthetic CXRs as is common in the literature; 
(iv) A two-stage data-centric training, where we first perform supervised pretraining with synthetic CXRs, followed by fine-tuning on real CXRs.
In scenarios (i)--(iii) models were initialized using ImageNet pretrained weights; in scenario (iv) models were pretrained from scratch using the synthetic data, then fine-tuned on real-only data.

Prior works have trained disease classifiers using a mix of real and synthetic data.
Khosravi et al.~\cite{LancetKhosravi2024} reported a $2.8\%$ increase in multi-label AUROC over baseline when using $10\times$ unconditional synthetic data supplementation; training with purely synthetic data ($2\times$ the number of real samples) matched the performance of the real-only baseline, but remained below that of the model trained with real and synthetic data. 
Ktena et al.~\cite{DeepmindKtena2024} reported a 5.2\% increase in multi-label AUROC over the real-only baseline when training a model on synthetic images only.
Our first set of experiments verified that supplementing real datasets with synthetic data enhances model performance and generalization, and that synthetic data alone does not match the mixed models. 
On average across five evaluation datasets, the model trained only on synthetic images increased AUROC by 2.1\% over the real-only baseline, while the model trained on real and synthetic images increased AUROC by 2.7\%.
Next, we enhanced current best practices by showing that synthetic images used during pretraining instead of as a data augmentation step further boosts classification performance across all datasets, leading to an average 6.5\% increase in AUROC. 
Our proposed pretraining strategy also increased robustness when fewer real samples were available for fine-tuning. 
By fine-tuning on less than $10k$ real samples, the synthetic-pretrained models matched or exceeded the AUROC of the real-only baseline trained on the full $66k$ real samples ($6\times$ more). 
When fine-tuning on $30k$ or more real samples, the synthetic-pretrained models exceeded the AUROC of the synthetic+real mix model trained on the combination of all available data.
None of the prior works investigated this low-data regime.

Next, motivated by prior work showing that intersectional subgroups often have compounded algorithmic biases~\cite{NatureSeyyedKalantari2021}, we conducted a detailed fairness analysis on intersectional subgroups of sex \& age and sex \& race/ethnicity.
Prior literature has assessed fairness of CXR classifiers using a variety of different metrics, where each metric has a different calculation and interpretation of fairness~\cite{chexclusion,NatureYang2024,DeepmindKtena2024}. 
Only one study~\cite{DeepmindKtena2024} probed the impact of synthetic data on fairness, and the analysis was limited to one metric (fairness gap) for subgroups defined by a single protected attribute (sex or race), on one out-of-distribution dataset.
In our experiments, for intersectional subgroups across four datasets, synthetic pretraining led to an average 16.0\% decrease in the fairness gap, and a 19.3\% decrease in the underdiagnosis gap.
Notably, average classification performance measured by AUROC increased across subgroups, irrespective of protected attributes.
Independent of the fairness metrics used, we showed that our synthetic data generation pipeline and data-centric synthetic pretraining approach enhance fairness in a generalized manner across several out-of-distribution datasets.

Our results should be interpreted with attention to some limitations.
First, we trained \modelname on one source dataset (MIMIC-CXR), and used only one radiographic view (PA), which limits the expressive power of the generative model.
The effects could be observed at inference time, where for example the model was not able to produce as many correct samples of Asian patients, likely owing to the much smaller training sample size for that population.
Second, when generating the synthetic dataset, we replicated the training dataset label distribution without changing disease prevalence. 
In future studies, oversampling of specific diseases with the low prevalence should be investigated for long-tailed disease classification.
Third, all classifiers are trained on datasets where disease labels were automatically extracted from radiology reports using the Chexpert labeler~\cite{irvin2019chexpert}.
Since this tool relies on rule-based natural language processing techniques, some of these labels may be incorrect, in ways that could compound with observed biases or model errors.
Lastly, while synthetic data holds great promise for advancing model development in healthcare, further research is needed to ensure privacy preservation and prevent patient re-identification that may occur through data memorization~\cite{Dar2025}.

Overall, our results demonstrate that high-quality, text and demographically controllable synthetic data can serve as a powerful complement to real datasets, improving both performance and fairness in medical imaging deep learning, especially when real data are imbalanced, scarce, or difficult to obtain.


\section{Methods}\label{sec4}
\subsection{\modelname Training}
\paragraph{Dataset} 
Our aim was to train a high-quality text-to-image diffusion model where the text prompt is derived from the radiology report impression section as well as patient demographics (sex, age, race/ethnicity).
We curated a paired image-text-demographics triplet dataset from the publicly available MIMIC-CXR~\cite{mimic-Johnson2019} data (version 2.0.0). 
MIMIC-CXR consists of $377k$ CXRs and corresponding text reports
obtained from $228k$ patients at the Beth Israel Deaconess Medical
Center (Boston, MA, United States).
We started with all available $377,095$ images from MIMIC-CXR. 
Our exclusion criteria included radiographic views other than postero-anterior, missing or non-descriptive impression section in the radiology report, missing sex or age information, and race/ethnicity not belonging to the Asian, Black, Hispanic, White categories. 
Following our exclusions, our dataset consisted of a total $75,639$ studies from MIMIC-CXR, with triplets of images, radiology reports, and demographic metadata (CONSORT diagram in Supplementary Figure~\ref{fig:consort}).
MIMIC-CXR is organized into ten folders (p10-p19). Based on prior studies~\cite{RoentgenBluethgen2024}, images belonging to p19 folder were used as a held-out test split ($7,584$ images). 
The images belonging to folders p10-p18 were divided into training ($66,760$ images) and validation ($1,295$ images) splits. 
No patients were shared across training, validation, and testing splits. 
Supplementary Table~\ref{supptab2} presents more details regarding the data composition of each split.
For each triplet of images, radiology reports, and demographic information we constructed a text prompt with the following structure: \texttt{<AGE> year old <RACE> <SEX>. <IMPRESSION>}. 
If the resulting prompt exceeded the 77 token limit of the CLIP text encoder~\cite{CLIP2021}, we abbreviated the impression using GPT-4~\cite{GPT4}, based on prior work demonstrating that GPT-4 is highly capable in summarizing radiology reports~\cite{RadAdapt2023,VanVeen2024}. 
For this summarization task, the GPT-4 prompt was: \textit{``Your task is to summarize this radiology report within 200 characters or less. Your response must be concise, truthful, and keep all relevant medical information.}". 
A total of $6,072$ impressions were summarized in this manner. 

\paragraph{Training}
The training procedure for \modelname followed best practices established in the original RoentGen paper~\cite{RoentgenBluethgen2024}.
We used Stable Diffusion~\cite{StableDiffusion2021}(version 2.1) as the underlying model with CLIP ViT-L/14 text encoder (77 token limit).
This was a newer version compared to the original RoentGen~\cite{RoentgenBluethgen2024} built upon Stable Diffusion version 1.4.
We jointly fine-tuned the text encoder and U-Net with learning rate $5 \times 10^{-5}$ for up to $60k$ optimization steps.  
We used AdamW optimizer and constant learning rate scheduler with 500 warmup steps. 
Training was performed on four A100 GPUs with 40GB vRAM each, allowing a batch size of 192.

\subsection{\modelname Evaluation and Checkpoint Selection}
For each model checkpoint, inference was performed using text prompts from the MIMIC-CXR validation set. 
Four synthetic images were generated per prompt using different random seeds to capture variability. 
Each synthetic image was paired with its corresponding real image, disease labels, and demographic attributes from the validation set. 
These pairs were used to compute a range of quality metrics, which guided the selection of the optimal model checkpoint for downstream experiments.

\paragraph{Text Prompt Alignment}
To evaluate how accurately \modelname adhered to the provided text prompts, we used pretrained classifiers from the Torch X-ray Vision (XRV) library~\cite{Cohen2022xrv}. 
For disease classification, we used the XRV DenseNet-121 model to compute the Area Under the Receiver Operating Characteristic curve (AUROC) for detecting Atelectasis, Cardiomegaly, Edema, Pneumothorax, and Pleural Effusion. 
For sex classification, we used the sex prediction model from~\cite{Glocker2023} and assessed performance using accuracy. For race classification, we used the XRV race model and evaluated accuracy. 
For age prediction, we used the XRV age model, reporting performance as the root mean square error (RMSE) in years. 
Since the images generated by \modelname were at a resolution of $512 \times 512$ pixels (higher than the $224 \times 224$ pixel input size expected by the XRV models) we downsampled the synthetic images using area-based interpolation prior to inference.

\paragraph{Real--Synthetic Image Similarity Assessment}
We evaluated the similarity between real and synthetic images by comparing both the raw images and their embeddings extracted from natural and biomedical image encoders, as detailed below.
The Fr\'{e}chet Inception Distance (FID) score~\cite{FID2017} quantifies the difference between the distributions of generated images and real ground truth images. 
Specifically, FID calculates the mean and covariance of embeddings derived from the deepest layer of the Inception v3 network~\cite{InceptionV3}, effectively measuring the similarity of high-level feature representations across the two image sets. 
A higher FID score indicates poorer generative performance, whereas a score of zero reflects a perfect match.

Recognizing that the conventional FID embeddings come from a network pretrained on natural images, we additionally generated embeddings using a model trained specifically on medical images. 
We used the encoder from the BioViL~\cite{BiovilBoecking2022}, a state-of-the-art CXR interpretation model to compute cosine similarity between embeddings of real images and their corresponding synthetic images generated from the same prompts. 
A cosine similarity score of 1 denotes identical embeddings, implying high fidelity in the synthetic images.

Furthermore, we calculated the multi-scale structural similarity index (MS-SSIM) between the distributions of real and synthetic images. 
MS-SSIM values range from 0 to 1, where 1 indicates identical images and 0 signifies no similarity. 
Importantly, neither extreme is desirable in this context, as MS-SSIM of 1 would reflect overfitting to the training data. Similarly, MS-SSIM of 0 would imply no similarity between generated images and their real counterparts, reflecting a severe degradation in the synthetic images’ fidelity to typical CXRs. 
We considered similarity scores below $0.25$ as indicative of failure to produce realistic CXRs.

\paragraph{Intra-Prompt Image Diversity}
For each text prompt in the validation set, four distinct synthetic images were generated with different random seeds. 
To quantify visual similarity among images produced from the same prompt, we computed average pairwise MS-SSIM for all four images. 
This per-prompt MS-SSIM reflects intra-prompt synthetic image similarity, with a value of 1 indicating identical images and lower values indicating greater variability.
The per-prompt MS-SSIM scores were then averaged across all prompts in the validation set, and the final mean MS-SSIM score reported as an overall measure of image-level intra-prompt diversity of the synthetic dataset.

Beyond image similarity, we also computed embedding similarity by measuring pairwise cosine distance between the BioViL embeddings of the four synthetic images generated per prompt.
A score of 1 indicates that the images are embedding-wise identical, while lower values reflect greater variability in semantic content as captured by the BioViL model.
These per-prompt cosine similarity scores were then averaged across all prompts in the validation set, and the final mean BioViL score reported as an overall measure of embedding-level intra-prompt diversity of the synthetic dataset.

For both mean MS-SSIM and mean BioViL, a score of $1$ indicates that all generated images are identical (model collapse), and lower scores indicate higher variability, which is desirable in the context of generative diversity.

\subsection{Synthetic Dataset Generation}
To generate a demographically balanced synthetic dataset, we constructed customized text prompts as follows. 
For each impression in the MIMIC-CXR training split, we created all possible combinations of sex (male, female) and race/ethnicity (White, Black, Asian, Hispanic), resulting in eight demographic variations per impression. 
For each variation, we randomly sampled an age within $\pm 5$ years of the original patient age. 
These demographic attributes were prepended to the original impression text to form the final prompt.
Image generation was performed using classifier-free guidance with a scale of 4.0 and the default noise scheduler, running inference for 75 steps to produce one image per prompt.
To ensure high fidelity to the specified demographic attributes and overall image quality, we implemented a quality control (QC) pipeline as follows. 
Each batch of generated images was evaluated using three pretrained XRV classifiers: one for sex, one for race, and one for age prediction. A synthetic image only passed quality control if the predicted sex and race exactly matched the specified value in the prompt and if the predicted age was within $\pm 7$ years of the age specified in the prompt.
The value of $\pm 7$ years was chosen equal to the measured standard deviation of the XRV age regression model when applied to real data.
Images that failed any of these criteria were discarded. 
All failed prompts were logged along with their failure mode, and a new round of image generation and QC was conducted on these prompts.
Due to compute constraints, each failed
prompt was re-generated a maximum of three times, then fully discarded.
Only high-quality images that passed our rigorous QC were retained for downstream analysis.

\subsection{Downstream Disease Classifier Training}
\paragraph{Datasets} 
The same MIMIC-CXR splits described above were used for the train, validation, and distribution testing of disease classifiers.
For out-of-distribution evaluation, we used four external chest radiography datasets: CheXpert~\cite{chexpert,chexpert-plus}, NIH~\cite{nih-Wang2017}, PadChest~\cite{padchest-Bustos2020} and VinDr~\cite{vindr-Nguyen2022}. 
CheXpert dataset consists of $224k$ CXRs from $65k$ patients at Stanford Health Care Hospital (Stanford, CA, United States).
NIH ChestX-ray dataset consists of $112k$ CXRs from $30k$ patients at the National Institutes of Health Clinical Center (Bethesda, MD, United States).
PadChest dataset consists of $160k$ CXRs from $67k$ patients at San Juan Hospital (Alicante, Spain).
VinDr-CXR dataset consists of $18k$ CXRs collected from the Hospital 108 and the Hanoi Medical University Hospital, two of the largest hospitals in Vietnam.
We included posteroanterior (PA) views only; in the case of VinDr, all images are frontal views but not specified whether PA or AP. 
For subgroup analysis, we used sex, age and race/ethnicity information available in MIMIC-CXR and CheXpert. 
We used sex and age information available in NIH and PadChest. We did not perform subgroup analysis on VinDr, since only a small number of patients had demographic information available. 
Supplementary Table~\ref{supptab2} summarizes the statistics of all image datasets.

\paragraph{Downstream Classifier Training}
We trained a 121-layer DenseNet models for multi-label classification task on the 14 CheXpert classes (13 diseases plus the `No Finding' label) with multi-label binary cross entropy loss. 
The 121-layer DenseNet was used as it produced the best results in prior studies~\cite{chexclusion,NatureSeyyedKalantari2021,NatureYang2024,RoentgenBluethgen2024}.
We followed identical procedures based on prior work for training CXR classification models~\cite{RoentgenBluethgen2024}.
Namely, all CXRs were center-cropped, resized to a resolution of 224 $\times$ 224 pixels, and normalized by the mean and standard deviation of the MIMIC-CXR train set.
All classifiers were trained using AdamW optimizer with cosine scheduler (initial learning rate 0.0001, weight decay 0.05), for a maximum of 100 epochs; we early stopped if validation loss did not improve over 20 epochs.

\paragraph{Strategies for Initializing Model Weights}
We conducted a comprehensive comparison of how synthetic data can be used to train disease classifiers, evaluating four strategies:
(i) Using only real CXRs;
(ii) Using only synthetic CXRs generated by \modelname;
(iii) Augmenting real CXRs with synthetic CXRs as is common in the literature;
(iv) A two-stage training, where we first perform supervised pretraining with synthetic CXRs, followed by fine-tuning on real CXRs.
In scenarios (i)--(iii) models were initialized using ImageNet pretrained weights; in scenario (iv) the model was trained from scratch. 
For models initialized with natural image (ImageNet) pretrained weights we used the available PyTorch weights.
For the synthetic pretraining strategy, we trained the DenseNet121 from scratch on all the available synthetic images  (initial learning rate 0.0001), and subsequently fine-tuned on real images  (initial learning rate 0.00005).
For each model, we saved the checkpoint with the highest average AUROC over 14 labels on the validation set.

\subsection{Disease Classifier Evaluation Metrics}
\paragraph{Classifier Performance}
To match the downstream disease classifiers that were trained only on PA images, we only used PA images in the test datasets. The only exception was VinDr, where all images were frontal views, but unspecified whether AP or PA. This may lead to reduced performance metrics on VinDr compared to the other datasets.
We evaluated classification performance on the set of eight labels common across datasets: Atelectasis, Cardiomegaly, Consolidation, Edema, Effusion, Pneumonia, Pneumothorax, No Finding.
We computed the area under the receiver-operating curve (AUROC) and area under the precision-recall curve (AUPRC).
Since this was a multi-label classification problem, binary AUROC was computed for each label and values were then macro-averaged across the eight labels of interest; similar for average AUPRC.
To obtain binarized predictions from the classifiers, we followed previous work in selecting the threshold that maximized the F1 score on the validation set~\cite{chexclusion,NatureSeyyedKalantari2021,NatureYang2024}. This threshold optimization procedure was conducted separately for each model and label.

\paragraph{Fairness Metrics}
For fairness analysis, we considered intersectional subgroups defined by pairs of demographic attributes. For sex \& race/ethnicity, the groups were: Asian males, Asian females, Black males, Black females, Hispanic males, Hispanic females, White males, and White females. 
For sex \& age, we defined age categories for both men and women: 18-40, 40-60, 60-80, and over 80 years old.
Supplementary Table~\ref{supptab3} details the subgroup composition of each dataset.

The first fairness metric we considered was AUROC parity, following prior work~\cite{DeepmindKtena2024}. 
To compute this metric, we first calculated the average AUROC within each demographic subgroup. 
The fairness gap was then defined as the absolute difference in AUROC between the highest- and lowest-performing subgroups.

Second, to assess model decision biases in underdiagnosed patients, we compared underdiagnosis rates across subpopulations. Following~\cite{NatureSeyyedKalantari2021, NatureYang2024}, we defined the underdiagnosis rate as the false positive rate (FPR) of the model prediction for the `No Finding' label at the levels of the subgroup.
We defined the underdiagnosis gap as the absolute difference in FPR between the best and worst-performing subgroups.

\paragraph{Statistical Analysis}
 
Continuous variables are reported as means with corresponding 95\% confidence intervals (CIs), estimated via 1,000 bootstrap resamples.
AUROC comparisons were conducted using the DeLong test, a nonparametric approach validated for evaluating differences between correlated receiver-operating characteristic curves~\cite{DeLong1988}.
Differences in AUPRC values were assessed using a permutation test, a nonparametric method suitable for evaluating performance metrics without distributional assumptions~\cite{Ernst2004}. Significance testing was performed with a type I error rate of 0.05. All processing was performed in Python (version 3.10.12), using the libraries numpy (version 1.25.2), scipy (version 1.16.1), pandas (version 2.3.0) and torchmetrics (version 1.7.3).


\subsubsection*{Data Availability}
The main data supporting the results in this study are available within the article and its Supplementary Information. 
The MIMIC-CXR dataset (version 2.0.0) is available on PhysioNet after a credentialing process (\url{https://doi.org/10.13026/C2JT1Q}). 
The CheXpert Plus dataset is available from Stanford AIMI (\url{https://aimi.stanford.edu/datasets/chexpert-plus}). 
The NIH Chest X-rays dataset is available from Kaggle (\url{https://www.kaggle.com/datasets/nih-chest-xrays/data}).
The PadChest dataset is available from BIMCV Medical Imaging Databank of the Valencia Region (\url{https://bimcv.cipf.es/bimcv-projects/padchest/}).
The VinDr-CXR dataset (version 1.0.0) is available on PhysioNet after a credentialing process (\url{https://doi.org/10.13026/3akn-b287}). 
Links to access the code, model weights, and synthetic dataset generated during this study are available at \url{https://github.com/StanfordMIMI/RoentGen-v2}.

\subsubsection*{Acknowledgments}
A.S.C. receives research support from NIH grants R01 HL167974, R01 HL169345, R01 AR077604, R01 EB002524, R01 AR079431, P41 EB027060; Advanced Research Projects Agency for Health (ARPA-H) contracts AY2AX000045 and 1AYSAX0000024-01; and the Medical Imaging and Data Resource Center (MIDRC), which is funded by the National Institute of Biomedical Imaging and Bioengineering (NIBIB) under contract 75N92020C00021 and through ARPA-H.
J.J. acknowledges co-funding by EU from EuroHPC Joint Undertaking programm under grant no. 101182737 (MINERVA) as well as co-funding by the Federal Ministry of Education and Research of Germany (BMBF) under the grant 16HPC117K (MINERVA), funding under grant no. 01IS24085C (OPENHAFM), and under the grant no. 01IS22094B (WestAI - AI Service Center West).
We gratefully acknowledge the Gauss Centre for Supercomputing e.V. for funding this work by providing computing time through the John von Neumann Institute for Computing (NIC) on the supercomputer JUWELS Booster at Jülich Supercomputing Centre (JSC), storage resources on JUST granted and operated by JSC and supported by Helmholtz Data Federation (HDF) and computing time granted by the JARA and JSC on the supercomputer JURECA at JSC.

\subsubsection*{Author Contributions}
S.L.M. and A.S.C. conceived and co-led the project and drafted the initial paper.
S.L.M., C.B. and P.C. trained RoentGen-v2 for CXR generation.
S.L.M. and C.B. trained and evaluated the supervised disease classifiers.
S.L.M. designed, implemented and conducted the experiments pertaining to synthetic data generation, pretraining with synthetic data, and fairness analysis.
B.P. and J.G. helped with demographic based classifiers.
J.B.D. and M.P. provided additional technical advice and assistance.
M.C. and J.J. provided engineering and computational support.
C.B., P.C., J.B.D., M.P., J.G., C.P.L. and A.S.C. helped revise the paper.
A.S.C. supervised the project.
C.P.L. and A.S.C. provided financial support for the project.
 
\nolinenumbers

\bibliographystyle{unsrt} 
\bibliography{main}     

\clearpage
\section*{\hspace{-0.5em}Supplementary Information}
\captionsetup[figure]{labelformat=default, labelsep=colon, name=Supplementary Figure}
\captionsetup[table]{labelformat=default, labelsep=colon, name=Supplementary Table}
\renewcommand{\thefigure}{\arabic{figure}}

\section*{Supplementary Results 1: \modelname Checkpoint Selection}
To select the best generative model checkpoint we computed several metrics to measure the quality of synthetic images across three distinct axes: (a) alignment with the provided findings text and demographics from the prompt, (b) similarity of generated images to real images, and (c) diversity among synthetic images generated from the same text prompt. 
Supplementary Table~\ref{supptab1} summarizes the synthetic image quality metrics at each model checkpoint.

\paragraph{(a) Text Prompt Alignment}
For each text prompt in the validation set we generated synthetic CXRs.
To evaluate how well synthetic CXRs depicted radiological findings that were included in the prompt, 
a pretrained classification model trained on real images (torch XRV~\cite{Cohen2022xrv}) was used to predict disease labels for the synthetic images.
Similarly, to evaluate how well synthetic CXRs depicted the demographic information from the prompt, separate XRV models, trained on real data, were used to predict sex, race and age. 
The disease labels and patient metadata of the original text prompt represented the ground truth labels for the respective tasks.
We report area under the receiver-operating characteristic curve (AUROC) for the diseases, accuracy for sex and race, and root mean square error (RMSE) for age predictions.
Since the XRV classifiers themselves are imperfect, we compute a reference baseline by evaluating the classifier performance on the real images from the validation set (see Supplementary Table~\ref{supptab1}, `real data' row).

The checkpoints at $7.5k$ training steps (equivalent to 21 epochs) and $10k$ training steps (equivalent to 28 epochs) achieved the highest average disease AUROC ($0.81$ compared to $0.88$ for the real data baseline). 
The original RoentGen~\cite{RoentgenBluethgen2024} model achieved  an average disease AUROC of $0.82$ (see Supplementary Table~\ref{supptab1}, `RoentGen' row), showing we maintain the disease fidelity while introducing metadata variables in the conditional image generation process.
For demographic attributes, sex and race accuracy for synthetic images were $100\%$ and $99\%$ respectively.
As a reference, the XRV sex and race accuracy in classifying the real images were $97\%$ and $95\%$ respectively.
This demonstrates the sex and race attributes were represented accurately by \modelname. 
The more challenging task of age prediction also had accurate results, with only a slightly larger root mean square error ($8.9$ yrs) compared to the real data ($7.1$ yrs).
We note that training past $15k$ optimization steps (equivalent to 43 epochs) led to decreased classification performance for disease, sex and race, likely due to model overfitting.
Overall, we observe near-perfect accuracy for demographics instruction following along with disease instruction following on par with the original RoentGen model.

\paragraph{(b) Real--Synthetic Image Similarity}
We used Fréchet Inception Distance (FID) score to compare the distribution of generated images with the distribution of the real ground truth images~\cite{FID2017}. 
Out of all model checkpoints, the $10k$ training steps (equivalent to 28 epochs) checkpoint achieved the lowest FID score (76.8), indicating the most ``realistic" image distribution.
By comparison, the FID score of images generated by RoentGen~\cite{RoentgenBluethgen2024} was 96.1.

While we aim to minimize the distance between the two image distributions, having individual synthetic images being exact replicas of real samples would be sub-optimal. 
To check for such model overfitting, we computed the image-level multi-scale structural similarity index (MS-SSIM) between generated images and their real counterparts, based on the same text prompt.
We similarly computed the embedding-level similarity by measuring the cosine similarity of image embeddings, extracted using the image encoder of a domain-specific foundation model (BioViL~\cite{BiovilBoecking2022}).
In the case of MS-SSIM, the highest observed score was $0.41$ at the $5k$ steps checkpoint (equivalent to 14 epochs), while the lowest observed score was $0.30$ at $30k$ and $40k$ steps (equivalent to 86 and 115 epochs), with all other checkpoints scoring in-between. 
For the BioViL metric, the highest observed score was $0.52$ at the $60k$ steps checkpoint (equivalent to 172 epochs), while the lowest observed score was $0.35$ at $5k$ steps (equivalent to 14 epochs), with all other checkpoints scoring in-between. 
We observed balanced similarity scores (MS-SSIM and BioViL) between synthetic and original images for all model checkpoints, indicating we do not overfit to the real samples.

\clearpage
\paragraph{(c) Intra-prompt Diversity of Generated Images}
To measure the diversity of synthetic images generated by \modelname, we generated multiple images using different random seeds for each text prompt in the validation set.
We computed pairwise similarity scores (MS-SSIM and cosine similarity of BioViL embeddings) between images generated from the same prompt.
For both metrics, a score of $1.00$ would indicate that the model collapsed and is generating identical samples, thus lower scores are desirable.
In the case of MS-SSIM, the highest observed score was $0.46$ at the $5k$ steps checkpoint (equivalent to 14 epochs), while the lowest observed score was $0.24$ at $30k$ and $40k$ steps (equivalent to 86 and 115 epochs), with all other checkpoints scoring in-between. 
For the BioViL metric, the highest observed score was $0.72$ at the $5k$ steps checkpoint (equivalent to 14 epochs), while the lowest observed score was $0.52$ at $20k$ steps (equivalent to 57 epochs), with all other checkpoints scoring in-between. 
Balanced intra-prompt similarity scores verify the ability of the model to generate distinct and varied images at multiple checkpoints.

\section*{Supplementary Tables}

\renewcommand{\thefootnote}{\arabic{footnote}}
\renewcommand{\thempfootnote}{\arabic{mpfootnote}}
\begin{sidewaystable}[htbp]
\caption{\modelname checkpoints: synthetic image quality metrics.}\label{supptab1}
\begin{tabular*}{\textheight}{@{\extracolsep\fill}lccccccccc|ccc|cc}
\toprule%
& \multicolumn{9}{@{}c@{}}{Text Prompt Alignment\footnotemark[2]}& 
\multicolumn{3}{@{}c@{}}{Real-Synthetic Similarity\footnotemark[3]}& 
\multicolumn{2}{@{}c@{}}{Intra-prompt Diversity\footnotemark[4]} 
\\\cmidrule{2-10}\cmidrule{11-13}\cmidrule{14-15}%
\makecell[l]{Checkpoint \\number of \\training steps \\(training epochs)} & 
\rotatebox{90}{Atelectasis} & 
\rotatebox{90}{Cardiomegaly} & 
\rotatebox{90}{Edema} & 
\rotatebox{90}{Effusion} & 
\rotatebox{90}{Pneumothorax} & 
\makecell{Average\\AUROC $\uparrow$} &
\makecell{Sex \\Acc. $\uparrow$}& 
\makecell{Race \\Acc. $\uparrow$}& 
\makecell{Age \\RMSE $\downarrow$} & 
FID $\downarrow$& 
MS-SSIM & 
BioViL & 
MS-SSIM $\downarrow$& 
BioViL $\downarrow$\\ 
\midrule
real data & 0.75 & 0.83 & 0.92 & 0.81 & 0.92 & 0.88 & 0.97 & 0.95 & 7.1 & 0.0 & 1.00 &  1.00 & - & - \\ 
\hline
RoentGen & 0.80 & 0.86 & 0.81 & 0.93 & 0.72 & 0.82 & - & - & - & 96.1 & 0.36 &  0.59 & 0.42 & 0.77 \\ 
\hline
5k (14 ep)& 0.72 & 0.81 & 0.84 & 0.73 & 0.87 & 0.79 & 1.00 & 0.98 & 9.4 & 77.2 & 0.41 &  0.35 & 0.46 & 0.72 \\ 
    
7.5k (21 ep)& 0.78 & 0.81 & 0.84 & 0.72 & 0.93 & 0.81 & 1.00 & 0.99 & 8.7 & 81.9 & 0.39 & 0.36 & 0.39 & 0.73 \\ 
    
10k (28 ep)& 0.76 & 0.77 & 0.82 & 0.78 & 0.90 & \textbf{0.81} & 1.00 & 0.98 & 8.9 & \textbf{76.8} & 0.37 & 0.41 & 0.37 & 0.66 \\ 

12.5k (35 ep)& 0.74 & 0.83 & 0.80 & 0.73 & 0.89 & 0.80 & 1.00 & 0.98 & 7.9 & 86.4 & 0.37 & 0.43 & 0.36 & 0.62 \\ 

15k (43 ep)& 0.76 & 0.80 & 0.79 & 0.73 & 0.89 & 0.80 & 0.99 & 0.98 & 8.4 & 87.4 & 0.34 & 0.44 & 0.31 & 0.57 \\ 

20k (57 ep)& 0.75 & 0.77 & 0.72 & 0.70 & 0.89 & 0.76 & 0.97 & 0.95 & 8.5 & 103.5 & 0.31 &  0.47 & 0.26 & 0.52 \\ 
    
30k (86 ep)& 0.70 & 0.78 & 0.71 & 0.66 & 0.89 & 0.75 & 0.98 & 0.96 & 8.7 & 101.3 & 0.30 & 0.48 & 0.24 & 0.54 \\ 
    
40k (115 ep)& 0.69 & 0.80 & 0.72 & 0.64 & 0.89 & 0.75 & 0.97 & 0.94 & 8.8 & 99.2 & 0.30 &  0.49 & 0.24 & 0.56 \\ 

50k (143 ep)& 0.72 & 0.81 & 0.73 & 0.65 & 0.90 & 0.76 & 0.98 & 0.95 & 8.9 & 97.9 & 0.31 &  0.51 & 0.26 & 0.60 \\ 

60k (172 ep)& 0.72 & 0.81 & 0.74 & 0.65 & 0.91 & 0.77 & 0.98 & 0.96 & 8.8 & 90.9 & 0.32 &  0.52 & 0.28 & 0.60 \\ 
\bottomrule
\end{tabular*}
\footnotetext[1]{Abbreviations: AUROC: area under the receiver-operating curve; FID: Frechet Inception Distance, calculated using embeddings from an ImageNet-pretrained InceptionV3 network; MS-SSIM: multi-scale structural similarity index; BioViL: indicates cosine similarity between image embeddings obtained from BioViL image encoder.}
\footnotetext[2]{Text Prompt Alignment Metrics. For individual disease labels, showing binary AUROC. For sex and race, showing classification accuracy. For age, showing root mean square error (RMSE).}
\footnotetext[3]{Real-Synthetic Similarity Metrics. For FID, lower scores are better. For MS-SSIM and BioViL, higher scores indicate more resemblance to real images; however, a score of 1.00 would indicate model overfitting, i.e. synthetic samples identical to their real counterparts.}
\footnotetext[4]{Intra-prompt Diversity Metrics. Lower scores indicate more diversity; a score of 1.00 would indicate model collapse, i.e. always identical synthetic samples.}
\end{sidewaystable}

\begin{sidewaystable}[htbp]
\caption{Description of chest radiography datasets.}\label{supptab2}%
\begin{tabular}{@{}lllllllll@{}}
\toprule
\textbf{Subgroup} & \textbf{Attribute} & \multicolumn{3}{c}{MIMIC-CXR} & CheXpert & NIH & PadChest & VinDr \\
\cmidrule(lr){3-5}
& Split & train & val & test & test & test & test & test \\
& Views & PA & PA & PA & PA & PA & PA & PA,AP\\
\midrule
Total & No.images & 66,760 & 1,295  & 7,584 &
20,531 & 47,075 & 59,036 & 3,000\\
\midrule
Sex 
& Female & 32,593 (48.8\%)  & 514 (39.7\%) & 3,670 (48.4\%) 
& 7,132 (34.7\%) & 20,872 (44.3\%) & 30,695 (52.0\%) & -\\
& Male   & 34,167 (51.2\%) & 781 (60.3\%) & 3,914 (51.6\%)
& 13,399 (65.3\%) & 26,203 (55.7\%) & 28,341 (48.0\%) & -\\
\midrule
Age & 0-18 & - & - & - & - & 1,922 (4.1\%) & 1,789 (3.0\%) & -\\
 & 18-40 & 9,599 (14.4\%) & 118 (9.1\%) & 1,186 (15.6\%) & 
 3,622 (17.6\%) & 12,684 (26.9\%) & 5,614 (9.5\%) & -\\
 & 40-60 & 24,167 (36.2\%) & 459 (35.4\%) & 2,606 (34.4\%) &
 7,179 (35.0\%) & 20,810 (44.2\%) & 16,666 (28.3\%) & -\\
 & 60-80 & 25,941 (38.8\%) & 574 (44.3\%) & 3,061 (40.4\%) & 
 7,722 (37.6\%) & 11,244 (23.9\%) & 22,682 (38.4\%) & -\\
 & 80+ & 7,053 (10.6\%) & 144 (11.2\%) & 731 (9.6\%) & 
 2,008 (9.8\%) & 415 (0.9\%) & 12,293 (20.8\%) & -\\
\midrule
\multirow{2}{*}{\shortstack{Race/\\Ethnicity \footnotemark[1]}} 
& Asian & 2,452 (3.6\%) & 14 (1.1\%) & 299 (3.9\%)
& 2,333 (11.4\%) & - & - & - \\
& Black & 13,685 (20.5\%) & 218 (16.8\%) & 1,500 (19.8\%)
& 1,174 (5.7\%) & - & - & - \\
& Hispanic & 5,521 (8.3\%) & 82 (6.3\%) & 700 (9.2\%)
& 2,401 (11.7\%) & - & - & - \\
& White & 45,102 (67.6\%) & 981 (75.8\%) & 5,085 (67.1\%)
& 11,395 (55.5\%) & - & - & - \\
\midrule
\multirow{2}{*}{\shortstack{Target\\Diseases \footnotemark[2]}} 
& No Finding & 34,028 (51.0\%) & 472 (36.4\%) & 3,783 (49.9\%) & 3,671 (17.9\%) 
& 27,379 (58.2\%) & 24,180 (41.0\%) & 2,051 (68.4\%)\\

& Atelectasis & 8,795 (13.2\%) & 210 (16.2\%) & 1,050 (13.8\%) & 4,531 (22.1\%) 
& 4,029 (8.6\%) & 2,477 (4.2\%) & 86 (2.9\%) \\

& Cardiomegaly & 7,074 (10.6\%) & 173 (13.4\%) & 831 (11.0\%) & 2,709 (13.2\%) 
& 1,122 (2.4\%) & 5,422 (9.2\%) & 309 (10.3\%)\\

& Consolidation & 2,209 (3.3\%) & 83 (6.4\%) & 209 (2.8\%) & 2,546 (12.4\%) 
& 1,059 (2.2\%) & 643 (1.1\%) & 96 (3.2\%) \\

& Edema & 4,332 (6.5\%) & 142 (11.0\%) & 504 (6.6\%) & 1,704 (8.3\%) 
& 181 (0.4\%) & 147 (0.3\%) & 0 (0.0\%) \\

& Effusion & 9,560 (14.3\%) & 267 (20.6\%) & 1,258 (16.6\%) & 6,542 (31.9\%) 
& 4,678 (9.9\%) & 2,032 (3.4\%) & 111 (3.7\%) \\

& Pneumonia & 8,806 (13.2\%) & 247 (19.1\%) & 976 (12.9\%) & 2,305 (11.2\%) 
& 438 (0.9\%) & 2,281 (3.9\%) & 246 (8.2\%) \\

& Pneumothorax & 1,696 (2.5\%) & 31 (2.4\%) & 197 (2.6\%) & 1,266 (6.2\%) 
& 2,386 (5.1\%) & 121 (0.2\%) & 18 (0.6\%) \\

\bottomrule
\end{tabular}
\end{sidewaystable}

\begin{table}[htbp]
\caption{Description of demographic subgroup composition of chest radiography datasets.}\label{supptab3}%
\centering
\begin{tabular}{@{}lllllllll@{}}
\toprule
    \textbf{Intersectional Subgroup} 
    & \multicolumn{3}{c}{MIMIC-CXR} 
    & CheXpert 
    & NIH 
    & PadChest \\
    
\cmidrule(lr){2-4}
    & train 
    & val 
    & test 
    & test 
    & test 
    & test \\
    
\midrule
    Total 
    & 66,760 
    & 1,295  
    & 7,584 
    & 20,531 
    & 47,075 
    & 59,036 \\
\midrule
    White Male 
    & $25,141$
    & 612
    & $2,880$
    & $7,572$
    & - 
    & - \\
    White Female 
    & $19,961$
    & 369
    & $2,205$
    & $3,550$
    & - 
    & - \\
    Black Male 
    & $5,366$
    & 105
    & 576
    & 653
    & - 
    & - \\
    Black Female 
    & $8,319$
    & 113
    & 924
    & 487
    & - 
    & - \\
    Hispanic Male 
    & $2,429$
    & 62
    & 325
    & $1,350$
    & - 
    & - \\
    Hispanic Female 
    & $3,092$
    & 20
    & 375
    & 992
    & - 
    & - \\
    Asian Male
    & $1,231$
    & 2
    & 133
    & $1,394$
    & - 
    & - \\
    Asian Female 
    & $1,221$
    & 12
    & 166
    & 886
    & - 
    & - \\
\midrule
    Male aged under 18 
    & 0
    & 0
    & 0
    & 0
    & $1,071$ 
    & 933 \\
    Female aged under 18
    & 0
    & 0
    & 0
    & 0
    & 851 
    & 856 \\
    Male aged $18-40$
    & $4,058$
    & 51
    & 469
    & $2,212$
    & $6,834$ 
    & $2,720$ \\
    Female aged $18-40$
    & $5,541$
    & 67
    & 717
    & $1,330$
    & $5,850$ 
    & $2,894$ \\
    Male aged $40-60$
    & $12,774$
    & 286
    & $1,370$
    & $4,558$
    & $11,166$ 
    & $7,447$ \\
    Female aged $40-60$
    & $11,393$
    & 173
    & $1,236$
    & $2,445$
    & $9,644$ 
    & $9,216$ \\
    Male aged $60-80$
    & $13,893$
    & 371
    & $1,681$
    & $5,087$
    & $6,880$ 
    & $11,343$ \\
    Female aged $60-80$
    & $12,048$
    & 203
    & $1,380$
    & $2,428$
    & $4,364$ 
    & $11,334$ \\
    Male aged over 80
    & $3,442$
    & 73
    & 394
    & $1,230$
    & 252 
    & $5,898$ \\
    Female aged over 80
    & $3,611$
    & 71
    & 337
    & 738
    & 163 
    & $6,395$ \\
\bottomrule
\end{tabular}
\footnotetext[1]{Note: NIH and PadChest datasets do not have race/ethnicity information available.}
\end{table}

\clearpage
\section*{Supplementary Figures}

\begin{figure}[ht]
    \centering
    \includegraphics[width=0.6\textwidth]{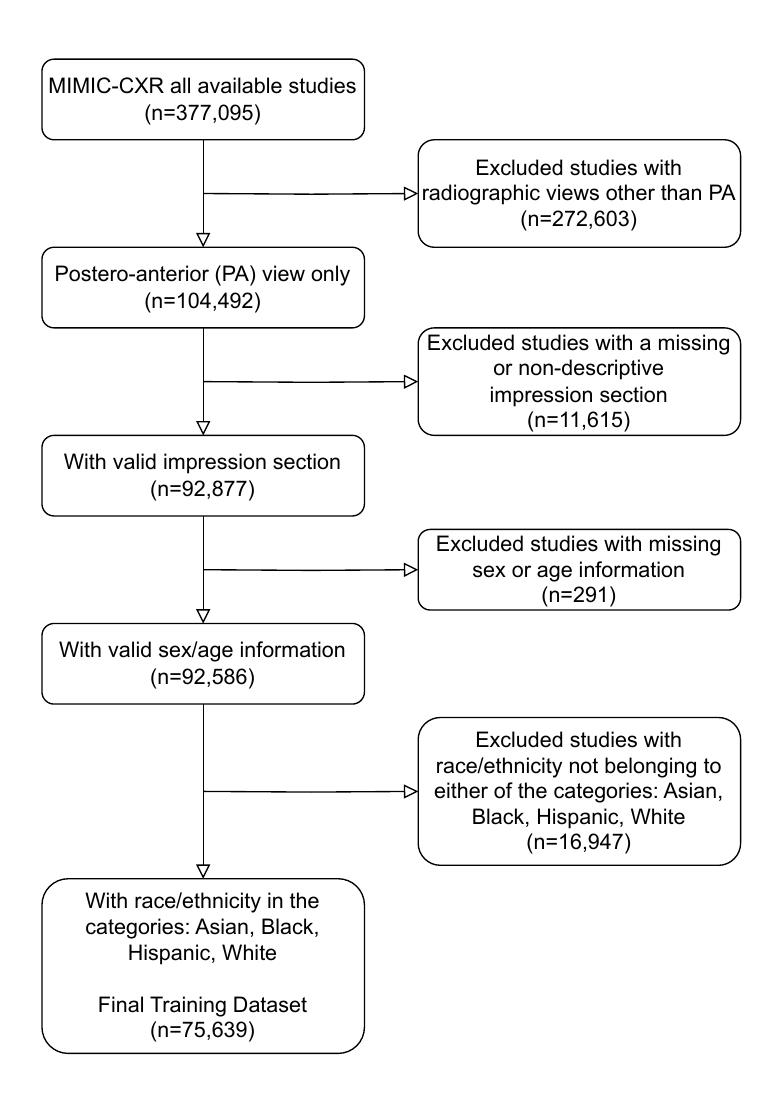}
    \caption{Selection criteria for the training dataset of the generative model.}
    \label{fig:consort}
\end{figure}

\begin{figure}[ht]
\centering
\includegraphics[width=0.62\textwidth]{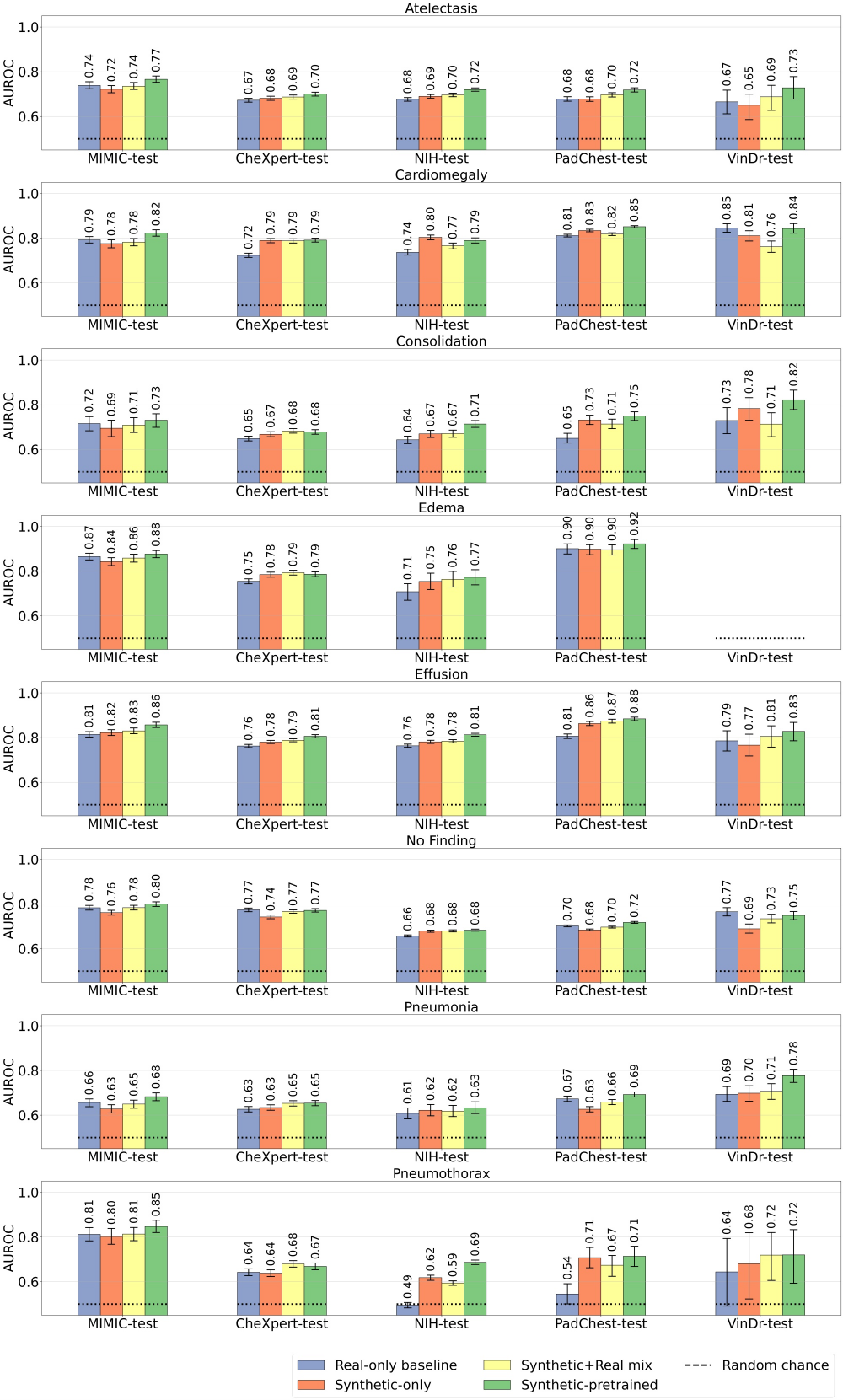}
\caption{Classification performance of models trained on all available real and/or synthetic data according to four strategies: 
(i) Real-only (baseline) model trained on $66k$ real CXRs, 
(ii) Synthetic-only model trained on $565k$ synthetic CXRs, 
(iii) Synthetic+Real mix model trained on combined $66k$ real and $565k$ synthetic CXRs, and 
(iv) Synthetic-pretrained model, which underwent supervised pretraining on $565k$ synthetic CXRs followed by fine-tuning on $66k$ real CXRs.
In scenarios (i)--(iii) models were initialized using ImageNet pretrained weights; in scenario (iv) the model was trained from scratch. 
Y-axis shows binary area under the receiver-operating curve (AUROC), with each panel focusing on one label. 
A random chance classifier would score 0.50 AUROC. 
For `Edema' label, there are no patients with positive label in VinDr-test. 
Each model is evaluated in-distribution on MIMIC-test and out-of-distribution on four external datasets.}\label{fig:suppfig1}
\end{figure}

\begin{figure}[ht]
\centering
\includegraphics[width=0.62\textwidth]{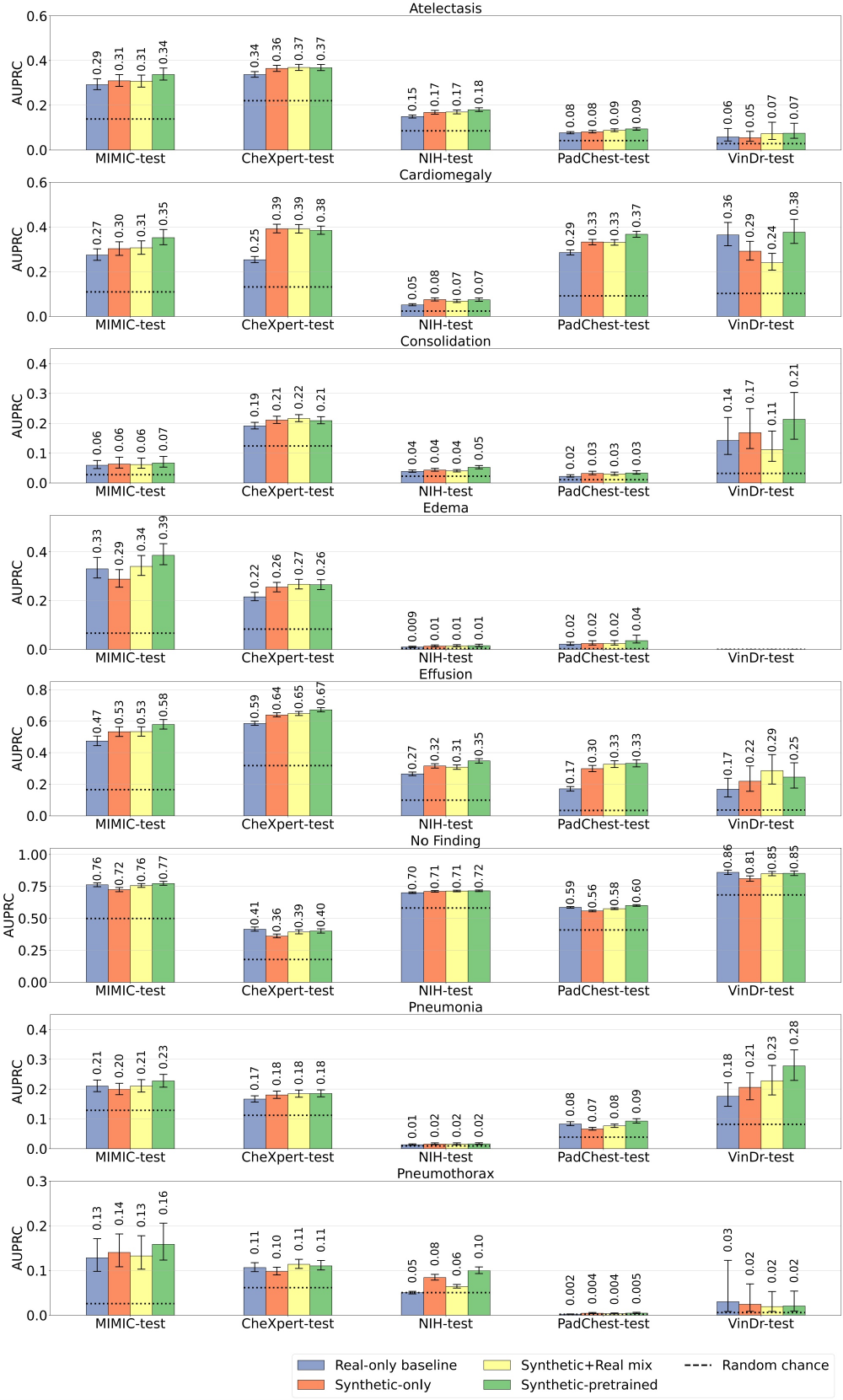}
\caption{Classification performance of models trained on all available real and/or synthetic data according to four strategies: 
(i) Real-only (baseline) model trained on $66k$ real CXRs, 
(ii) Synthetic-only model trained on $565k$ synthetic CXRs, 
(iii) Synthetic+Real mix model trained on combined $66k$ real and $565k$ synthetic CXRs, and 
(iv) Synthetic-pretrained model, which underwent supervised pretraining on $565k$ synthetic CXRs followed by fine-tuning on $66k$ real CXRs.
In scenarios (i)--(iii) models were initialized using ImageNet pretrained weights; in scenario (iv) the model was trained from scratch. 
Y-axis shows binary area under the precision-recall curve (AUPRC), with each panel focusing on one label. 
A random chance classifier would score AUPRC equal to the label prevalence in the test dataset, hence the large differences in scores between labels and datasets. 
Each model is evaluated in-distribution on MIMIC-test and out-of-distribution on four external datasets.
For `Edema' label, there are no patients with positive label in VinDr-test; the disease prevalence is 0.004 in NIH-test, and 0.002 in PadChest-test. 
For `Pneumonia' label, the disease prevalence in NIH-test is 0.009.
For `Pneumothorax' label, the disease prevalence is 0.002 in PadChest-test, and 0.006 in VinDr-test.}\label{fig:suppfig2}
\end{figure}

\clearpage
\section*{Supplementary Note 1: Disease-specific Classification Metrics}
Supplementary Figure~\ref{fig:suppfig1} shows the disease-specific AUROC metrics across all models and datasets.
The trend observed in the macro-averaged multi-label AUROC was reflected in the binary classification AUROCs of individual imaging findings. 
Generally, the synthetic-pretrained model achieved the highest AUROC scores across datasets and diseases, with only a few exceptions.
For Cardiomegaly on VinDr, there was no significant difference between the real-only baseline the synthetic-pretrained model (p-value$=0.80$).
For Consolidation on MIMIC, there was no significant difference between the real-only baseline the synthetic-pretrained model (p-value$=0.18$).
Lastly, for the No Finding label on CheXpert, there was no significant difference between the real-only baseline the synthetic-pretrained model (p-value$=0.43$), and on VinDr the real-only model achieved higher AUROC than the synthetic-pretrained model (p-value$=0.02$).

Supplementary Figure~\ref{fig:suppfig2} shows the disease-specific AUPRC metrics across all models and datasets.
Across the disease labels of Atelectasis, Cardiomegaly, Consolidation, Edema, Pleural Effusion, Pneumonia and Pneumothorax, the synthetic-pretrained model achieved the best binary AUPRC. 
For the `No Finding' label, there was no significant difference between the proposed synthetic pretraining strategy and the baseline model.

\section*{Supplementary Note 2: Details of Pretrained CXR Classifiers}
\begin{itemize}
    \item Torch X-ray Vision library~\cite{Cohen2022xrv}: XRV version 1.3.5
    
    \item Disease classification model (XRV): \\ \texttt{xrv.models.DenseNet(weights="densenet121-res224-all")}
    
    \item Race classification model (XRV): \\ \texttt{xrv.baseline\_models.emory\_hiti.RaceModel()} 
    
    \item Age prediction model (XRV): \\
    \texttt{xrv.baseline\_models.riken.AgeModel()} 

    \item Sex classification model from~\cite{Glocker2023}: \\ \texttt{SexModelResNet.load\_from\_checkpoint("sex\_model\_chexpert\_resnet\_all.ckpt")} with weights from \url{https://github.com/biomedia-mira/chexploration/tree/main/prediction}

    \item Downstream disease classification models initialized with natural image (ImageNet) pretrained weights used the PyTorch weights \\ \texttt{DenseNet121\_Weights.IMAGENET1K\_V1}. 
\end{itemize}


\end{document}